%% file: main.tex
\documentclass{article}
\PassOptionsToPackage{table}{xcolor}
\usepackage{guidednav_preprint,times}
\usepackage[T1]{fontenc}
\usepackage[utf8]{inputenc}
\usepackage{amsmath,amssymb,bm}
\usepackage{graphicx}
\usepackage{booktabs,multirow}
\usepackage[table]{xcolor}
\usepackage{url}
\usepackage{xurl}
\usepackage{hyperref}
\usepackage{tabularx}
\usepackage{capt-of}
\usepackage{subcaption}
\usepackage{placeins}
\usepackage{float}
\usepackage{microtype}
\hypersetup{hidelinks}

\graphicspath{{figures/}}
\hypersetup{
  pdftitle={GuidedNav: Shaping Vision-Language Navigation Representations through Pre-Action Attention and Spatial Guidance},
  pdfauthor={Jingyu Guo, Jiaxin Huang, Chengxing Lin, Xiaoyu Luo, Cheng Wen, Shanshan Ye, Tingjin Chu, Shaoli Huang, Mingming Gong},
  pdfsubject={Vision-and-language navigation},
  pdfkeywords={GuidedNav, vision-and-language navigation, attention guidance, spatial guidance}
}

\newif\ifdraftnotes
\draftnotesfalse

\newif\ifincludeappendix
\includeappendixtrue
\title{GuidedNav: Shaping Vision-Language Navigation Representations through Pre-Action Attention and Spatial Guidance}

\author{%
  Jingyu Guo\textsuperscript{1,2}\quad
  Jiaxin Huang\textsuperscript{2,3,\textdagger}\quad
  Chengxing Lin\textsuperscript{2,4}\quad
  Xiaoyu Luo\textsuperscript{2,5}\\[3pt]
  Cheng Wen\textsuperscript{2}\quad
  Shanshan Ye\textsuperscript{6}\quad
  Tingjin Chu\textsuperscript{1}\quad
  Shaoli Huang\textsuperscript{2,*}\quad
  Mingming Gong\textsuperscript{1,3,*}
}
\newcommand{\authoraffiliations}{%
  \textsuperscript{1}University of Melbourne\qquad
  \textsuperscript{2}AgiBot\\
  \textsuperscript{3}Mohamed bin Zayed University of Artificial Intelligence\\
  \textsuperscript{4}University of Electronic Science and Technology of China\\
  \textsuperscript{5}University of Macau\qquad
  \textsuperscript{6}University of Technology Sydney
}
\newcommand{\correspondence}{\textsuperscript{\textdagger}Project leader.\quad\textsuperscript{*}Corresponding authors.}

\begin{document}
\maketitle

\begin{center}
  \small
  \textbf{Project page:} \url{https://jingyu198.github.io/GuidedNav/}
\end{center}

\begin{abstract}
\input{abstract}
\end{abstract}

\input{introduction}
\input{related_work}

\input{method}
\input{experiments}
\input{conclusion}

\FloatBarrier
\section*{AI Use Statement}
We used large language models to assist with manuscript drafting, translation, language refinement, and code implementation.
We also used generative AI tools to assist with schematic illustrations, and employed gpt-oss-120b and GLM-4.5V in the annotation pipeline described in Section~\ref{sec:data_engine}.
The authors are responsible for verifying all AI-assisted content and for the accuracy and integrity of the final manuscript, implementation, and reported results.
\begingroup
\raggedright
\setlength{\bibsep}{2pt}
\bibliography{references}
\bibliographystyle{iclr2027_conference}
\endgroup

\ifincludeappendix
  \clearpage
  \appendix
  \input{appendix}
\fi
\end{document}

%% file: abstract.tex
Vision-and-language navigation (VLN) requires an agent to track progress through an extended instruction, identify landmarks relevant to its current objective, and understand the surrounding spatial structure to act.
End-to-end navigation policies typically rely on action supervision to implicitly learn these capabilities.
Without explicit guidance on the evidence underlying each decision, however, policies can overfit to correlations predictive of training actions rather than learn the task-relevant grounding and geometry needed for generalization.
Our insight is to guide the representations that connect instructions to actions: identifying what remains to be done determines which visual evidence matters, while understanding the geometry around that evidence informs how to move.
We introduce \textbf{GuidedNav}, a unified framework that explicitly shapes these representations through training-time guidance.
Subtask and landmark guidance steer native pre-action attention toward the active instruction segment and relevant visual regions, grounding action prediction in the current objective.
Spatial guidance aligns the policy's internal representations with geometric features that capture long-horizon context and local spatial structure.
Together, these objectives strengthen navigation capabilities while avoiding the inference overhead of intermediate textual reasoning and online geometric encoding.
We also introduce \textbf{Guided-R2R} and \textbf{Guided-RxR}, two annotation-enriched navigation datasets that support more interpretable VLN research.
GuidedNav establishes state-of-the-art performance among the compared monocular RGB policies, improving success rates by 8.1\% and 13.1\% relative to the strongest prior results on R2R-CE and RxR-CE val-unseen, respectively, while achieving faster inference.
Extensive experiments further demonstrate the effectiveness and real-world transferability of our approach.

%% file: introduction.tex
\suppressfloats[t]
\begin{figure}[t]
\centering
\includegraphics[width=0.95\linewidth]{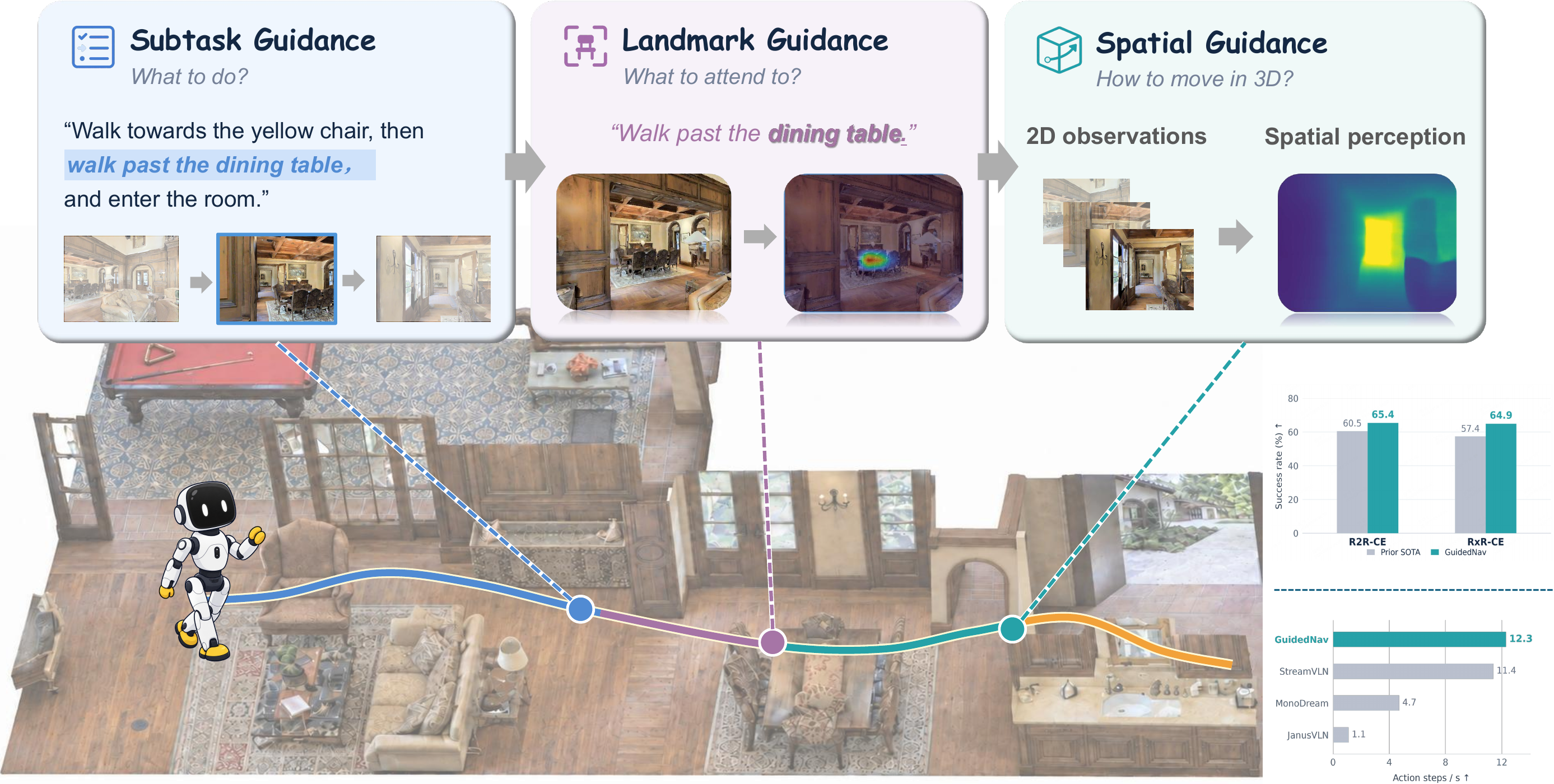}
\caption{\textbf{Guidance Designs in GuidedNav.}
Subtask guidance focuses native pre-action attention on the active instruction span (left); landmark guidance directs it toward task-relevant visual regions (middle); spatial guidance shapes representations of the surrounding 3D structure using historical and nearby observations (right).
Together, these training-time objectives connect instruction progress, visual grounding, and geometry to direct action prediction along the illustrated route.}
\label{fig:concept_overview}
\end{figure}

\section{Introduction}
\label{sec:introduction}
Consider the instruction ``Walk towards the yellow chair, then walk past the dining table, and enter the room'' (Figure~\ref{fig:concept_overview}).
Moving forward may be appropriate at each stage, although the active instruction span and relevant landmark differ.
The action label alone therefore does not specify which evidence should support a decision.
This ambiguity matters in vision-and-language navigation (VLN), where an agent follows extended instructions through changing visual observations~\citep{anderson2018vln,ku2020rxr}.
Recent vision-language policies predict actions directly from instructions and RGB video~\citep{zhang2024navid,zhang2025uninavid,cheng2025navila}, but their behavioral supervision only indirectly constrains the task and visual information used to act.

Grounding the current objective is only part of the problem: the policy must also relate 2D observations to the 3D layout through which it moves.
Historical views provide broader spatial context, while nearby observations reveal local structure relevant to the next movement.
Existing work strengthens these capabilities through progress reasoning, region-level grounding, and geometric memory~\citep{wang2026progressthink,guo2026awarevln,li2023kerm,zeng2026janusvln}.
Intermediate text generation and online geometric encoding can, however, add inference cost.
Auxiliary reasoning and future-state prediction provide additional training signals~\citep{wang2025auxthink,zhang2026futurenav}, motivating a complementary question: at the moment of action generation, does the policy attend to the active instruction span and its visual referents, and encode the spatial information needed to proceed?

We introduce \textbf{GuidedNav}, a unified framework that strengthens navigation representations through training-time auxiliary guidance.
The position immediately preceding the first action token provides a natural supervision point: its hidden state predicts the action sequence's beginning, while its native attention gathers evidence from instructions and observations.
Subtask and landmark guidance direct this attention toward the active instruction span and associated image regions, grounding action prediction in the current objective.
Spatial guidance aligns current-frame representations with long-horizon geometric targets and supervises local temporal fusion with geometric information from nearby views.
At inference, the policy retains the learned fusion but discards the geometric teacher and auxiliary projectors, enabling faster, direct action prediction without intermediate reasoning.

We introduce \textbf{Guided-R2R} and \textbf{Guided-RxR}, two annotation-enriched datasets linking navigation steps to active subtask spans and landmark masks.
GuidedNav improves success rates by 8.1\% and 13.1\% relative to the strongest prior models on R2R-CE and RxR-CE, respectively, while achieving a $10.7\times$ per-action inference speedup.
Our contributions are threefold:
\begin{itemize}
    \item We propose GuidedNav, a unified framework that explicitly guides navigation representations through subtask attention, landmark grounding, and spatial supervision, strengthening direct action prediction without intermediate reasoning or an online geometric encoder.
    \item We introduce Guided-R2R and Guided-RxR through a landmark-centric data engine, linking navigation steps to active subtask spans and landmark masks to support representation learning and more interpretable VLN research.
    \item We achieve state-of-the-art navigation performance with faster inference. Extensive benchmark evaluations, component ablations, representation analyses, and real-robot experiments validate the effectiveness of our guidance and its transfer to real-world environments.
\end{itemize}

%% file: related_work.tex
\section{Related Work}
\label{sec:related_work}
\noindent\textbf{Vision-and-language navigation.}
VLN has evolved from navigation graphs to continuous action execution~\citep{anderson2018vln,ku2020rxr,krantz2020vlnce}.
Graph- and map-based methods combine grounding with planning~\citep{chen2022duet,an2023bevbert,wang2023gridmm,an2024etpnav}, while pretrained vision-language policies enable RGB-based action prediction and robot deployment~\citep{zhang2024navid,zhang2025uninavid,cheng2025navila}.
GuidedNav focuses on the task and spatial representations supporting these decisions.

\smallskip\noindent\textbf{Instruction execution and visual grounding.}
Progress monitoring and auxiliary reasoning extend action supervision~\citep{ma2019selfmonitoring,zhu2020auxrn}; sub-instruction methods and Landmark-RxR align instructions with trajectories and landmarks~\citep{hong2020subinstruction,he2021landmarkrxr}.
Recent policies generate reasoning or progress descriptions~\citep{zhou2024navgpt,lin2025navcot,wang2026progressthink,guo2026awarevln}, or learn from auxiliary reasoning and future-state prediction~\citep{wang2025auxthink,zhang2026futurenav}.
KERM and GELA strengthen region/entity grounding~\citep{li2023kerm,cui2023gela}; HINT aligns gradient-based explanations with human annotations~\citep{selvaraju2019hint}, and GuidedVLA specializes action-decoder attention heads~\citep{jia2026guidedvla}.
GuidedNav instead supervises native pre-action attention to the active subtask span and landmarks, without adding specialized heads or generating intermediate text.

\smallskip\noindent\textbf{Spatial representations for navigation.}
Maps organize observations for spatial reasoning~\citep{an2024etpnav,an2023bevbert,wang2023gridmm}; StreamVLN models temporal context through streaming attention~\citep{wei2026streamvln}, while JanusVLN retains VGGT-based geometric memory at inference~\citep{zeng2026janusvln,wang2025vggt}.
Spatial Forcing transfers geometry through training-time representation alignment~\citep{li2026spatialforcing}.
GuidedNav extends this principle to sequential navigation with history-conditioned targets and a local objective that supervises temporal fusion, incorporating both temporal scales without online teachers.

%% file: method.tex
\section{Method}
\label{sec:method}

\subsection{Problem Formulation}
\label{sec:formulation}
We consider vision-and-language navigation in continuous environments~\citep{krantz2020vlnce}.
At step $t$, the agent predicts a short action sequence from instruction $I$, sampled historical RGB observations $H_t$, and current observation $o_t$:
\begin{equation}
    A_t \sim \pi_\theta(\cdot \mid I,H_t,o_t).
    \label{eq:policy}
\end{equation}
Actions comprise forward movement, left/right turns, and stop; observations update after execution until stopping.
Here $A_t=(a_{t,1},\ldots,a_{t,m_t})$ contains up to four primitive actions ($m_t\leq4$), terminating early when stop is predicted.
We learn representations connecting the active instruction segment, landmarks, and spatial context for action prediction.

\subsection{GuidedNav Framework}
\label{sec:framework}
\begin{figure}[t]
\centering
\includegraphics[width=\linewidth]{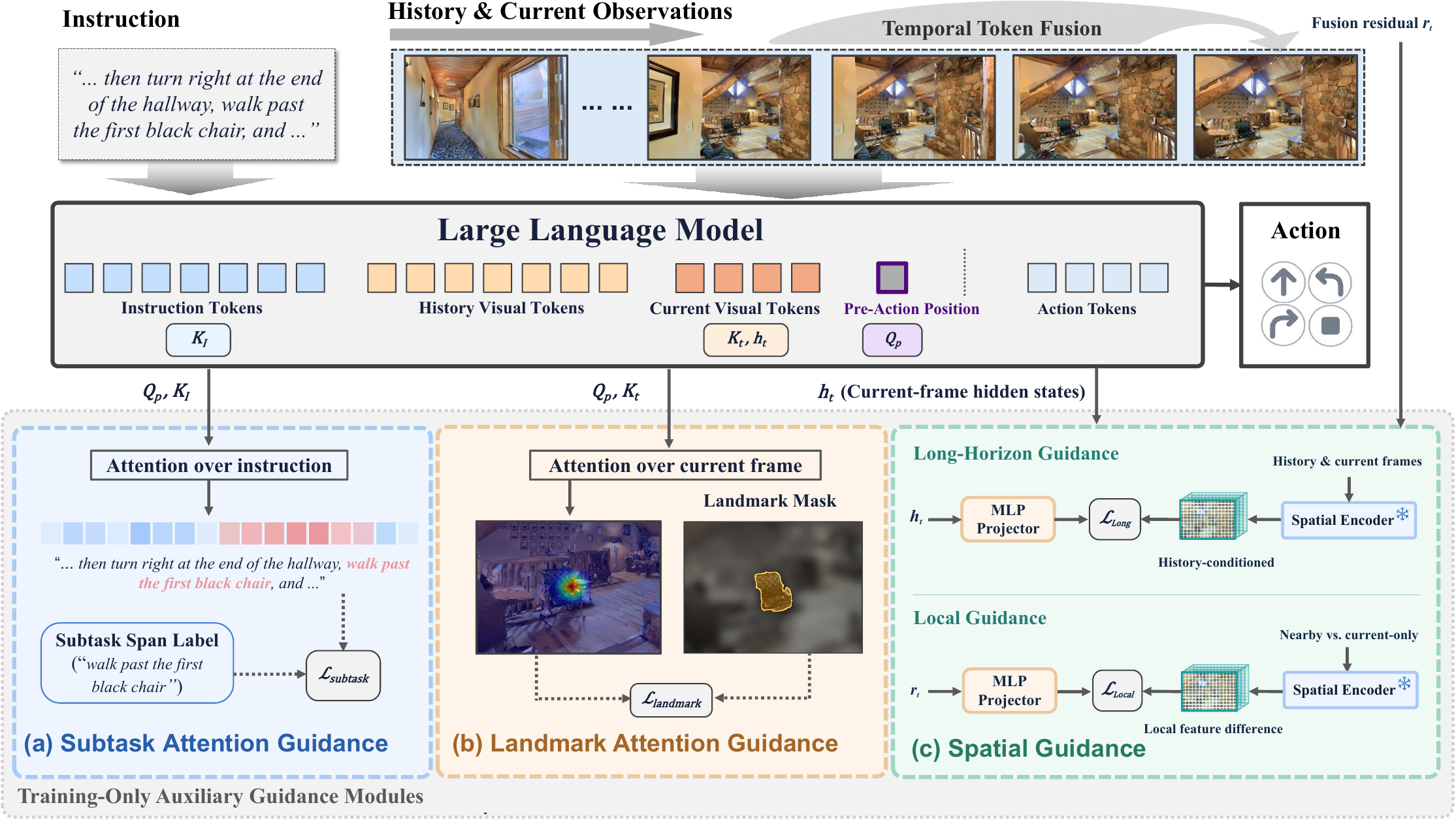}
\caption{\textbf{Architecture of GuidedNav.}
Subtask and landmark guidance supervise native pre-action attention toward the active instruction span and relevant visual regions.
Spatial guidance aligns current-frame representations with history-conditioned geometric features and supervises local fusion using geometric feature differences from nearby observations.
All auxiliary objectives and the frozen spatial encoder are used only during training.}
\label{fig:overview}
\end{figure}

GuidedNav augments an autoregressive vision-language policy with training-time guidance (Figure~\ref{fig:overview}): subtask and landmark objectives supervise native language and visual attention, while spatial guidance combines long-horizon geometry with nearby observations.
Inference predicts actions directly.

\paragraph{Native pre-action attention.}
We read attention at the existing position immediately preceding the first action token, which predicts the beginning of the action sequence.
Following the notation in Figure~\ref{fig:overview}, $Q_p$ denotes the query at this pre-action position, while $K_I$ and $K_t$ denote keys of instruction tokens and current-frame visual tokens.
For a selected set of attention heads $\mathcal{B}$, we compute
\begin{equation}
    \alpha_D = \frac{1}{|\mathcal{B}|}
    \sum_{b\in\mathcal{B}}
    \operatorname{softmax}\!\left(
        \frac{Q_p^{(b)}(K_D^{(b)})^\top}{\sqrt{d_b}}
    \right), \qquad D\in\{I,t\},
    \label{eq:native_attention}
\end{equation}
where $d_b$ is the head dimension.
The query and keys retain the backbone's positional encoding, and softmax is applied separately within each token group before averaging across heads.
Both domains exclude action tokens; the readout introduces no query tokens or attention layers.

\subsubsection{Subtask Attention Guidance}
\label{sec:subtask_guidance}
An extended instruction contains several subtasks, only some of which are relevant to the current decision.
Let $\mathcal{S}_t$ denote the token indices of the active subtask span, obtained from the annotations in Section~\ref{sec:data_engine}.
We encourage the pre-action position to attend to this span through
\begin{equation}
    \mathcal{L}_{\mathrm{Subtask}}
    = -c_t\log\left(\sum_{i\in\mathcal{S}_t}\alpha_{I,i}\right),
    \label{eq:subtask_loss}
\end{equation}
where $c_t$ balances the contribution of subtask intervals, preventing long intervals from dominating the span objective (Appendix~\ref{app:supervision_weights}).
This objective increases attention to the active span while allowing the policy to distribute attention among its tokens.
Subtask labels provide training supervision and are not required as policy inputs.

\subsubsection{Landmark Attention Guidance}
\label{sec:landmark_guidance}
To connect instruction execution with visual evidence, we guide attention toward landmarks associated with the active subtask.
Accepted landmark masks are combined and area-pooled onto the current visual-token grid, yielding patch occupancies $m_{t,j}\in[0,1]$.
We maximize the attention mass supported by these regions:
\begin{equation}
    \mathcal{L}_{\mathrm{Landmark}}
    = -\log\left(\sum_{j=1}^{N_t}m_{t,j}\alpha_{t,j}\right),
    \label{eq:landmark_loss}
\end{equation}
where $N_t$ is the number of current-frame visual tokens.
The mask weights preserve fractional patch coverage rather than forming a normalized target distribution.
Together, the two attention objectives connect the active instruction segment and its visual referents to the same action-predicting position.

\subsubsection{Spatial Guidance}
\label{sec:spatial_guidance}
A single RGB view provides incomplete depth and occlusion cues; sampled history and nearby views offer complementary context.
We use a frozen VGGT teacher for long-horizon alignment and local fusion supervision (Figure~\ref{fig:spatial}).

\paragraph{Long-horizon spatial guidance.}
Frozen VGGT~\citep{wang2025vggt} jointly encodes sampled history and the current frame.
Its final aggregated patch features $f_t^{\mathrm{long}}$ are aligned to the policy's current-frame grid after adding positional embeddings; Appendix~\ref{app:feature_alignment} specifies the preprocessing.
An MLP $P_{\mathrm{long}}$ aligns current-frame states $h_{t,j}$ from hidden-state index 27 ($h_t$ in Figure~\ref{fig:overview}) with these history-conditioned targets:
\begin{equation}
    \mathcal{L}_{\mathrm{Long}}
    = \frac{1}{N_t}\sum_{j=1}^{N_t}
    \left[1-\cos\left(P_{\mathrm{long}}(h_{t,j}),
    f_{t,j}^{\mathrm{long}}\right)\right].
    \label{eq:long_spatial}
\end{equation}
History supplies context to both networks; alignment covers only the current frame's full patch grid, independently of landmark masks.

\paragraph{Temporal token fusion.}
Before the language model, we refine current visual-encoder tokens $z_{t,j}$ using up to three immediately preceding observations.
After aligning token-grid resolutions, cosine similarity with a spatial-distance penalty selects mutual best matches $\mathcal{M}_t(j)$, retaining at most 25\% of the grid per past frame (Appendix~\ref{app:fusion_matching}): 

\begin{equation}
    r_{t,j} = \frac{\sum_{(u,i)\in\mathcal{M}_t(j)}w_{t-u}(z_{u,i}-z_{t,j})}
    {\sum_{(u,i)\in\mathcal{M}_t(j)}w_{t-u}},
    \qquad \widetilde z_{t,j}=z_{t,j}+g\,r_{t,j}.
    \label{eq:temporal_fusion}
\end{equation}
Here $w_{t-u}$ weights temporal distance, $g$ is a learned bounded scalar gate, and unmatched tokens have $r_{t,j}=0$.
Fusion preserves token count and spatial arrangement.

\paragraph{Local spatial guidance.}
To supervise fusion, we compare current-frame teacher features from a local temporal window, $f_t^{\mathrm{local}}$, and the current frame alone, $f_t^{\mathrm{single}}$.
Aligned to the current patch grid, their normalized difference $\Delta f_t$ captures the teacher-feature change induced by nearby views (Equation~\ref{eq:local_target}, Appendix~\ref{app:feature_alignment}).
A separate projector $P_{\mathrm{local}}$ aligns the pre-gate residual with this target over patches with valid matches, denoted by $\mathcal{J}_t$:
\begin{equation}
    \mathcal{L}_{\mathrm{Local}}
    = \frac{1}{|\mathcal{J}_t|}\sum_{j\in\mathcal{J}_t}
    \left[1-\cos\left(P_{\mathrm{local}}(r_{t,j}),
    \Delta f_{t,j}\right)\right].
    \label{eq:local_spatial}
\end{equation}
We omit this term without valid matches. Long-horizon guidance supervises current representations; local guidance supervises fusion residuals. Both use only current and past observations.

\subsection{Landmark-Centric Data Engine}
\label{sec:data_engine}
\begin{figure}[t]
\centering
\includegraphics[width=\linewidth]{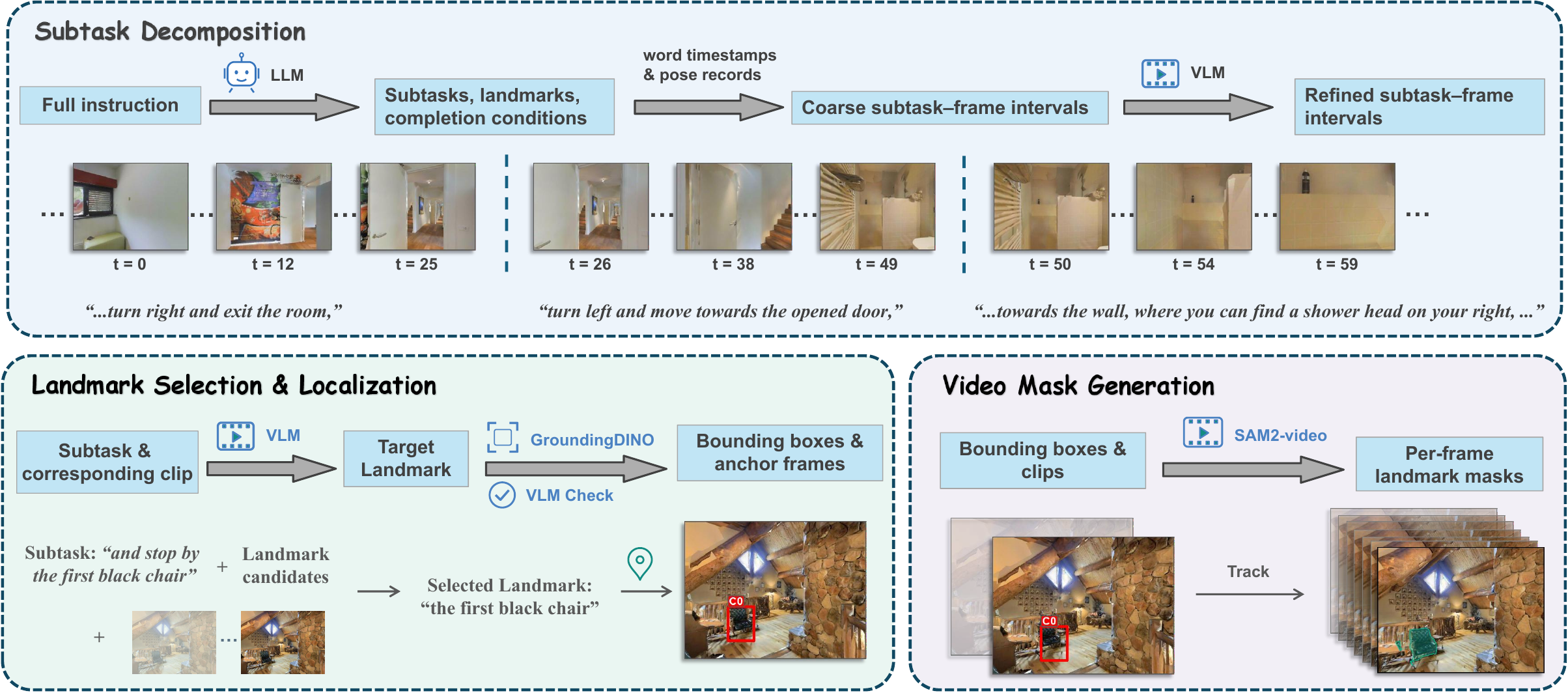}
\caption{\textbf{Landmark-Centric Data Engine for Guided-R2R and Guided-RxR.}
The pipeline combines subtask decomposition and temporal alignment, landmark selection and localization, and per-frame mask generation to produce annotations for navigation training.}
\label{fig:annotation_pipeline}
\end{figure}

We construct Guided-R2R and Guided-RxR by augmenting R2R and English RxR training instructions with step-aligned subtask spans and landmark masks.
Figure~\ref{fig:annotation_pipeline} summarizes the offline pipeline: subtask decomposition and temporal alignment, landmark selection and localization, and per-frame mask generation.
R2R spans come from FGR2R~\citep{hong2020subinstruction}, mapped to continuous trajectories using recorded poses; RxR spans are extracted by gpt-oss-120b and temporally refined by GLM-4.5V.
For RxR, word timestamps and guide poses provide coarse subtask--frame intervals before visual refinement.
Within each aligned clip, GLM-4.5V selects task-relevant landmarks, verifies GroundingDINO boxes~\citep{liu2024groundingdino}, and selects anchor frames.
SAM2~\citep{ravi2025sam2} propagates accepted box prompts within each subtask interval to produce per-frame landmark masks.
Masks are filtered by area, detection consistency, and temporal support.
Unreliable annotations are excluded from the corresponding auxiliary loss, while original observations, instructions, and action targets are retained.
These annotations supervise training and are not required at inference; Appendix~\ref{app:data_engine} provides further construction details.

\subsection{Training and Inference}
\label{sec:training_inference}
GuidedNav combines action-token cross-entropy with attention guidance and spatial guidance, where $\mathcal{L}_{\mathrm{Spatial}}=\lambda_{\mathrm{long}}\mathcal{L}_{\mathrm{Long}}+\lambda_{\mathrm{local}}\mathcal{L}_{\mathrm{Local}}$:
\begin{equation}
    \mathcal{L}=\mathcal{L}_{\mathrm{Action}}+\eta(s)\big[
    \lambda_{\mathrm{sub}}\mathcal{L}_{\mathrm{Subtask}}
    +\lambda_{\mathrm{lm}}\mathcal{L}_{\mathrm{Landmark}}
    +\mathcal{L}_{\mathrm{Spatial}}\big].
    \label{eq:total_objective}
\end{equation}
where $\eta(s)$ gradually introduces auxiliary supervision at the beginning of training.
Each auxiliary term is averaged over its valid training examples, and teacher features are detached from optimization.
At inference, GuidedNav retains the learned policy and temporal fusion, while discarding the geometry teacher and auxiliary projectors.
It consumes RGB observations and instructions without requiring subtask labels, landmark masks, or intermediate textual reasoning.

%% file: experiments.tex

\section{Experiments}
\label{sec:experiments}

\subsection{Experimental Setup}
\label{sec:setup}

\paragraph{Datasets and three-stage training.}
We evaluate on R2R-CE and RxR-CE val-unseen~\citep{anderson2018vln,ku2020rxr,krantz2020vlnce}, using English guide instructions for RxR.
Training starts from Qwen3-VL~\citep{bai2025qwen3vl} and proceeds in three stages.
Stage 1 fully fine-tunes the model on approximately 0.53M R2R and 1.90M RxR expert examples, together with 0.79M visual instruction-tuning (VQA) examples from LLaVA-Video~\citep{zhang2025llavavideo} and VSI~\citep{yang2025cambrians}.
Stage 2 continues full fine-tuning on 2.97M DAgger-collected navigation examples~\citep{ross2011dagger} mixed with VQA data.
Stage 3 applies LoRA fine-tuning to Guided-R2R and Guided-RxR.
Throughout the experiments, \textbf{Base Model} denotes the same Stage-2 checkpoint followed by one epoch of LoRA fine-tuning on the original R2R and RxR data without our designed guidance modules, providing the comparison for GuidedNav.

\paragraph{Training configuration.}
Each stage runs for one epoch with global batch size 256, AdamW, cosine learning-rate decay, 3\% warmup, and weight decay 0.01.
Stages 1 and 2 update the language model, multimodal projector, and vision tower, with peak language-model learning rates of $2\times10^{-5}$ and $1\times10^{-5}$, respectively; the projector and vision tower use $5\times10^{-6}$ in both stages.
Stage 3 uses rank-32 LoRA with scaling factor 16 and a peak learning rate of $2\times10^{-4}$ for both Base Model and GuidedNav.
Training uses BF16 and DeepSpeed ZeRO-2. See Appendix~\ref{app:training_recipe} for more details. 

\paragraph{Metrics and evaluation protocol.}
We report navigation error (NE), oracle success rate (OSR), success rate (SR), and success weighted by path length (SPL), together with normalized dynamic time warping (nDTW) on RxR~\citep{ilharco2019ndtw}.
NE is measured in meters and the remaining metrics in percentages.
Evaluation uses greedy decoding with up to eight context images and four-action execution chunks.
Each primitive action moves forward by 0.25\,m or rotates by $15^\circ$, with a maximum of 500 actions per episode.
Table~\ref{tab:simulator_settings} in Appendix~\ref{app:implementation} lists the remaining simulator settings.

\input{tables/main_results}

\begin{figure}[t]
\centering
\captionsetup{
    font=small,
    skip=6pt,
    justification=raggedright,
    singlelinecheck=false
}

\begin{minipage}[t]{0.43\linewidth}
    \vspace{0pt}
    \centering
    \includegraphics[width=\linewidth]{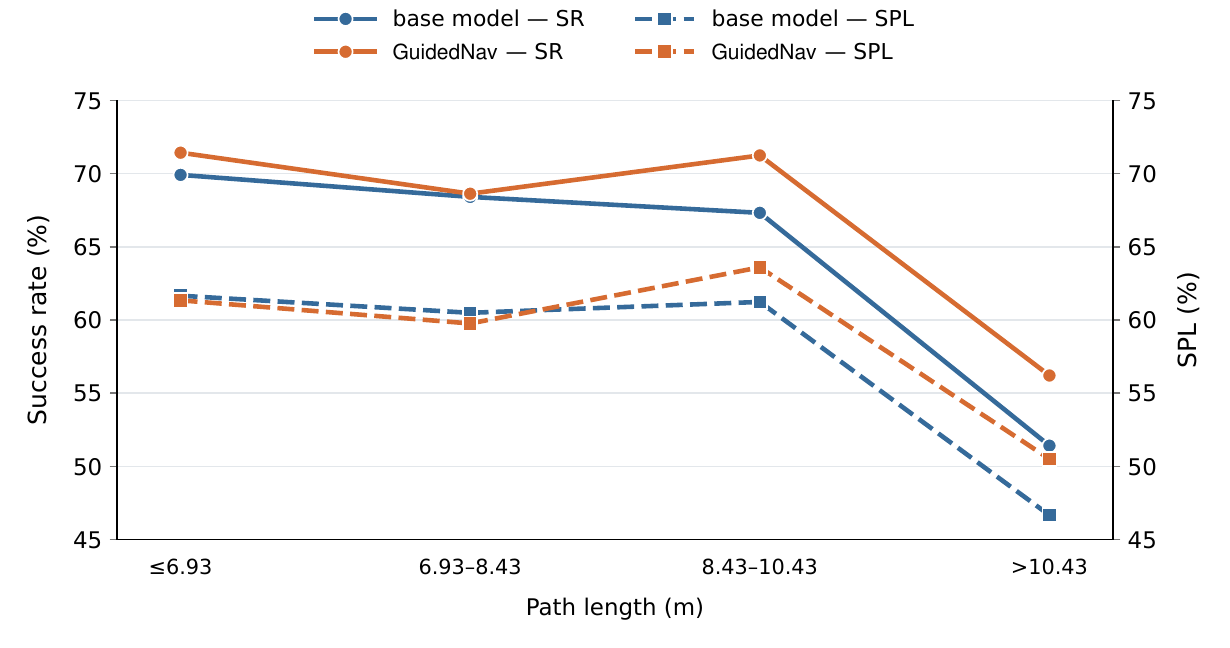}
    \captionof{figure}{\textbf{Performance across Path-Length Groups}
    on R2R-CE val-unseen. }
    \label{fig:path_length}
\end{minipage}
\hfill
\begin{minipage}[t]{0.54\linewidth}
    \vspace{0pt}
    \centering
    \small
    \setlength{\tabcolsep}{4pt}
    \renewcommand{\arraystretch}{1.25}
    \captionof{table}{\textbf{Inference Efficiency}
    on a single NVIDIA RTX 5090.
    Latency is amortized per executed primitive action;
    throughput is measured in action steps per second.}
    \label{tab:inference_speed}

    \resizebox{\linewidth}{!}{%
    \begin{tabular}{@{}l|rrr@{}}
        \toprule
        Method & Params (B)
        & ms/step$\downarrow$ & steps/s$\uparrow$ \\
        \midrule
        StreamVLN~\citep{wei2026streamvln}
        & 8.03 & 87 & 11.4 \\
        JanusVLN~\citep{zeng2026janusvln}
        & 9.31 & 864 & 1.1 \\
        MonoDream~\citep{wang2026monodream}
        & 2.00 & 211 & 4.7 \\
        \midrule
        \rowcolor{blue!7}
        \textbf{GuidedNav (Ours)}
        & 8.77 & \textbf{81} & \textbf{12.3} \\
        \bottomrule
    \end{tabular}%
    }

\end{minipage}
\end{figure}

\subsection{Main Results}
\label{sec:main_results}

\paragraph{Results on VLN-CE benchmarks.}
Table~\ref{tab:main_results} shows that GuidedNav advances the performance frontier for monocular RGB navigation.
Relative to the strongest prior result for each metric, it improves SR by 8.1\% on R2R-CE and 13.1\% on RxR-CE, alongside SPL gains of 4.0\% and 9.9\%.
The SR comparisons use JanusVLN~\citep{zeng2026janusvln} on R2R and NavFoM~\citep{zhang2026navfom} on RxR, corresponding to gains of 4.9 and 7.5 percentage points.
Thus the improvement extends beyond reaching more goals to completing routes more efficiently.
On RxR, nDTW also improves by 10.5\% over the best prior RGB result, indicating closer adherence to the instructed route rather than endpoint success alone.
This joint improvement is consistent with guidance linking the current objective to visual evidence and spatial context throughout navigation.
Additional comparisons and trajectory visualizations appear in Appendices~\ref{app:additional_results} and~\ref{app:rollouts}.

\paragraph{Performance across route lengths.}
Figure~\ref{fig:path_length} compares GuidedNav with Base Model across four path-length groups.
SR changes little and SPL is slightly lower in the two shorter groups, whereas both metrics improve in the 8.43--10.43\,m and $>10.43$\,m groups.
The benefit is therefore concentrated where the policy must maintain the instruction-to-scene connection over more decisions; on shorter routes, the small SPL decrease suggests that guidance does not uniformly improve path efficiency.

\subsection{Inference Efficiency}
\label{sec:inference_efficiency}
Table~\ref{tab:inference_speed} reports measurements on a single NVIDIA RTX 5090, amortized over the primitive actions executed from each prediction.
GuidedNav achieves 12.3 steps/s at 81\,ms per step, yielding $10.7\times$ and $2.6\times$ per-step speedups over JanusVLN and MonoDream, respectively, and 6.9\% lower latency than StreamVLN.
The central efficiency advantage is architectural: geometric learning is moved into training, so deployment retains temporal fusion but discards the teacher and auxiliary projectors.
This allows the policy to benefit from spatial guidance without repeatedly encoding geometry or generating intermediate reasoning.

\subsection{Ablation Study}
\label{sec:ablation_study}

\input{tables/component_ablations}

The upper block of Table~\ref{tab:ablations} shows complementary effects of attention and spatial guidance under the same Stage-3 protocol.
Subtask, landmark, and spatial guidance improve SR over Base Model by 3.6, 3.3, and 3.1 percentage points, respectively.
Attention guidance gives the largest OSR gains, suggesting better access to the evidence needed to reach the goal vicinity, whereas spatial guidance also improves SPL by 1.1 points.
Combining the three signals improves SR by 6.1 points and SPL by 3.0 points over Base Model, outperforming each individual module on both metrics.

The lower block isolates the progressive spatial design.
Beyond long-horizon guidance, gated fusion adds 0.2 SR and 0.1 SPL points; supervising the local residual adds another 0.4 and 0.5 points.
The larger incremental SPL gain from local supervision suggests that the content of the fused temporal information matters more than fusion alone.
Appendix~\ref{app:module_analysis} compares alternative supervision strategies: direct attention guidance gives the highest SR for both subtask and landmark learning, supporting supervision at the action-predicting position without requiring textual intermediate outputs.

\subsection{Representation Analysis}
\label{sec:representation_analysis}

\input{tables/attn_vis}

\begin{figure}[t]
\centering
\includegraphics[width=0.8\linewidth]{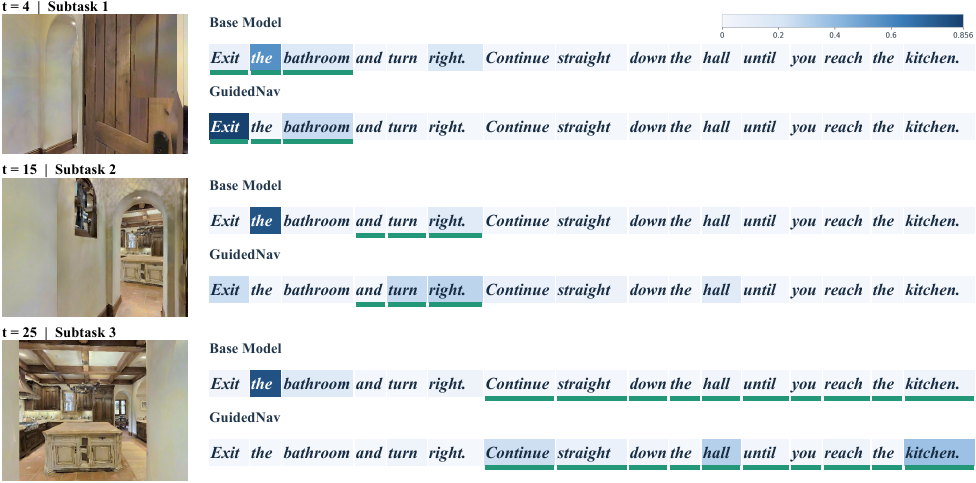}
\caption{\textbf{Subtask Attention across Navigation Steps.}
We compare the Base Model and GuidedNav at sampled steps.
Green underlines mark the GT active subtask, while darker
blue indicates higher native pre-action attention.}
\label{fig:subtask_attention}
\end{figure}

\begin{figure}[t]
\centering
\begin{minipage}[t]{0.48\linewidth}
    \vspace{0pt}
    \centering
    \includegraphics[
        width=\linewidth,
        trim=0 198 0 0,
        clip
    ]{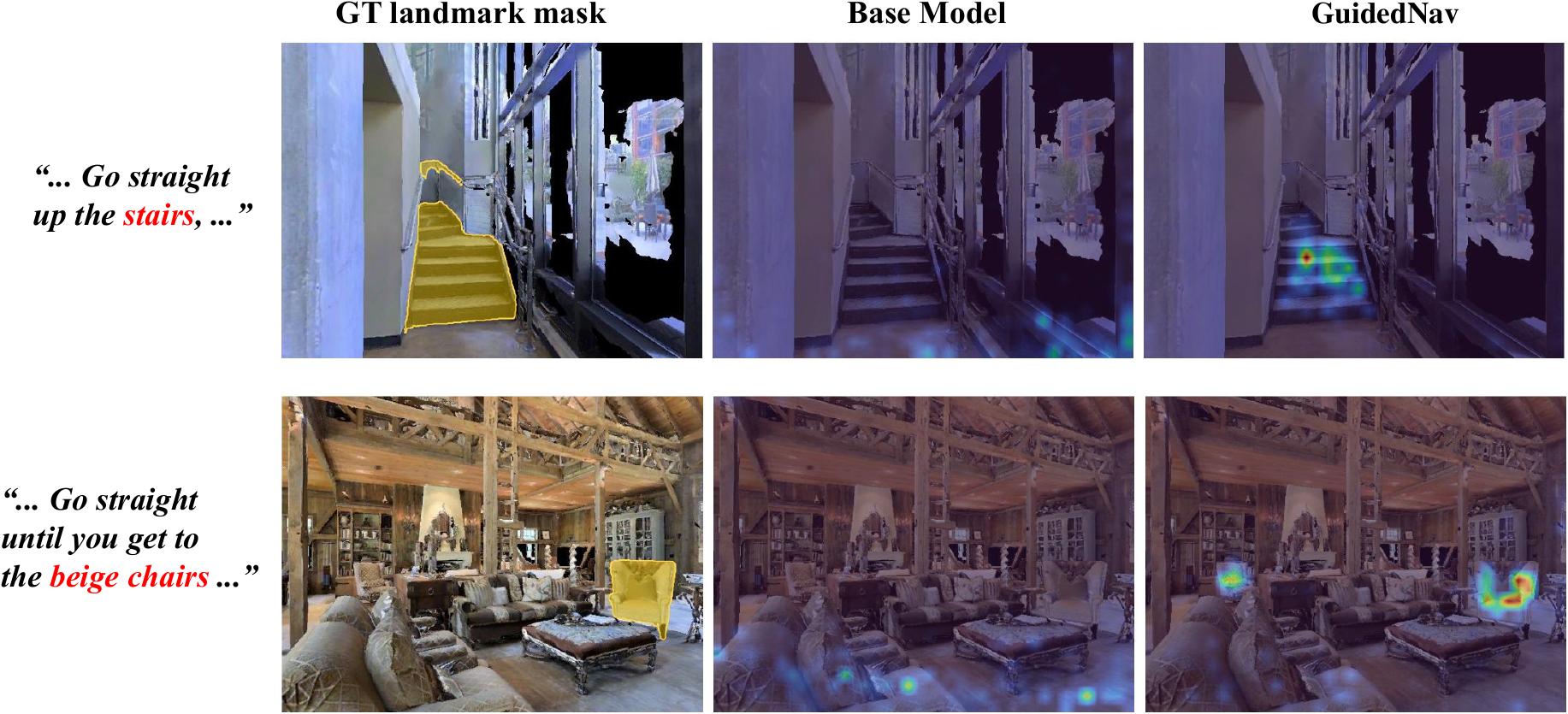}
    \caption{\textbf{Landmark Grounding at the Pre-Action Position.}
    Reference masks (left), Base Model attention (middle), and
    GuidedNav attention (right) for stairs and chairs.}
    \label{fig:landmark_main}
\end{minipage}
\hfill
\begin{minipage}[t]{0.48\linewidth}
    \vspace{0pt}
    \centering
    \includegraphics[width=\linewidth]{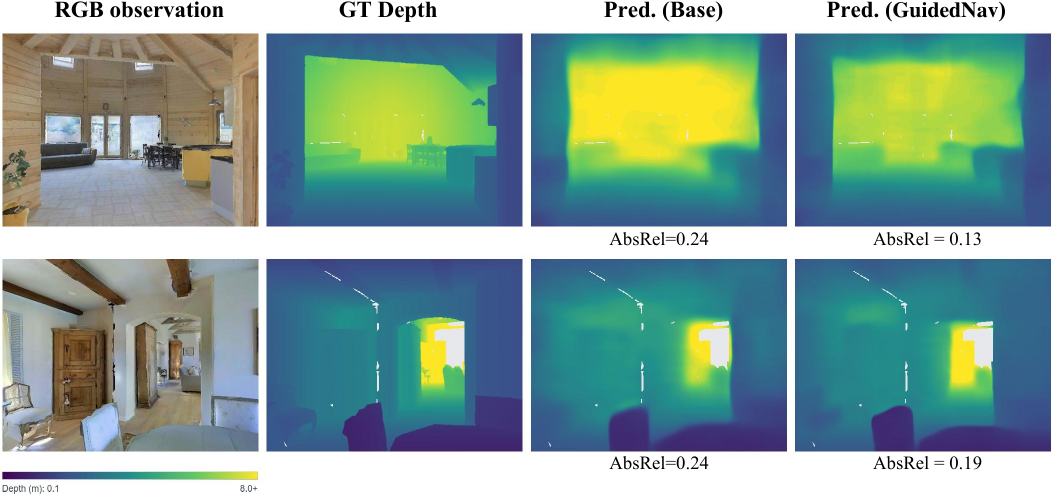}
    \caption{\textbf{Depth Decoded from Frozen Navigation Representations.}
    RGB, reference depth, and Base Model/GuidedNav predictions,
    with per-example AbsRel and depth in meters.
    }
    \label{fig:depth_comparison}
\end{minipage}
\end{figure}

\paragraph{Task and landmark localization.}
We visualize native attention from the position immediately preceding the first action token.
Attention is normalized separately over instruction tokens and current-frame visual tokens, then averaged across the selected heads.
The resulting scores highlight words in the instruction and form heatmaps on the visual-token grid.
Table~\ref{tab:attention_localization} shows gains of 21.8 and 22.9 points in subtask AUPRC and AUROC, alongside landmark PG and EBPG improvements of 33.9 and 29.2 points.
The subtask gains indicate better separation of the active instruction span from completed or upcoming subtasks.
The landmark gains extend beyond placing the attention peak inside the target: more attention mass is also assigned to the annotated regions, suggesting more consistent selection of task-relevant visual evidence.
Figure~\ref{fig:subtask_attention} illustrates how this emphasis changes with instruction progress, while Figure~\ref{fig:landmark_main} shows stronger localization of the relevant landmarks.
Together, these observations suggest that guidance helps organize evidence around the current navigation objective rather than merely producing a fixed preference for salient words or objects.
These maps describe relative attention within each token group; they do not directly measure the causal contribution of individual tokens to an action.
Further details appear in Appendix~\ref{app:module_analysis}.

\paragraph{Spatial information in frozen representations.}
We assess spatial information by fitting depth probes to frozen navigation features, allowing the learned representations to be examined.
For visualization, a separate convolutional head reads current-frame features and predict a depth map.
This head is used only for analysis and does not participate in action prediction.
GuidedNav reduces probe AbsRel by 24.0\% and RMSE by 22.0\%, while improving near--far ranking accuracy by 5.8 points (Table~\ref{tab:spatial_probe}).
The joint improvement matters because depth errors and ranking accuracy capture complementary properties: the former assess metric reconstruction, whereas the latter tests whether the representation preserves relative spatial ordering.
Figure~\ref{fig:depth_comparison} provides a qualitative view of the recovered scene structure, with lower per-example depth errors for GuidedNav in both displayed cases.
These results suggest that spatial guidance makes geometry more accessible within the policy itself, even after the geometric teacher is removed.
Details are shown in Appendix~\ref{app:spatial_probes}.

\FloatBarrier
\subsection{Real-World Navigation}
\label{sec:real_world}
\begin{figure}[H]
\centering
\includegraphics[width=0.95\linewidth]{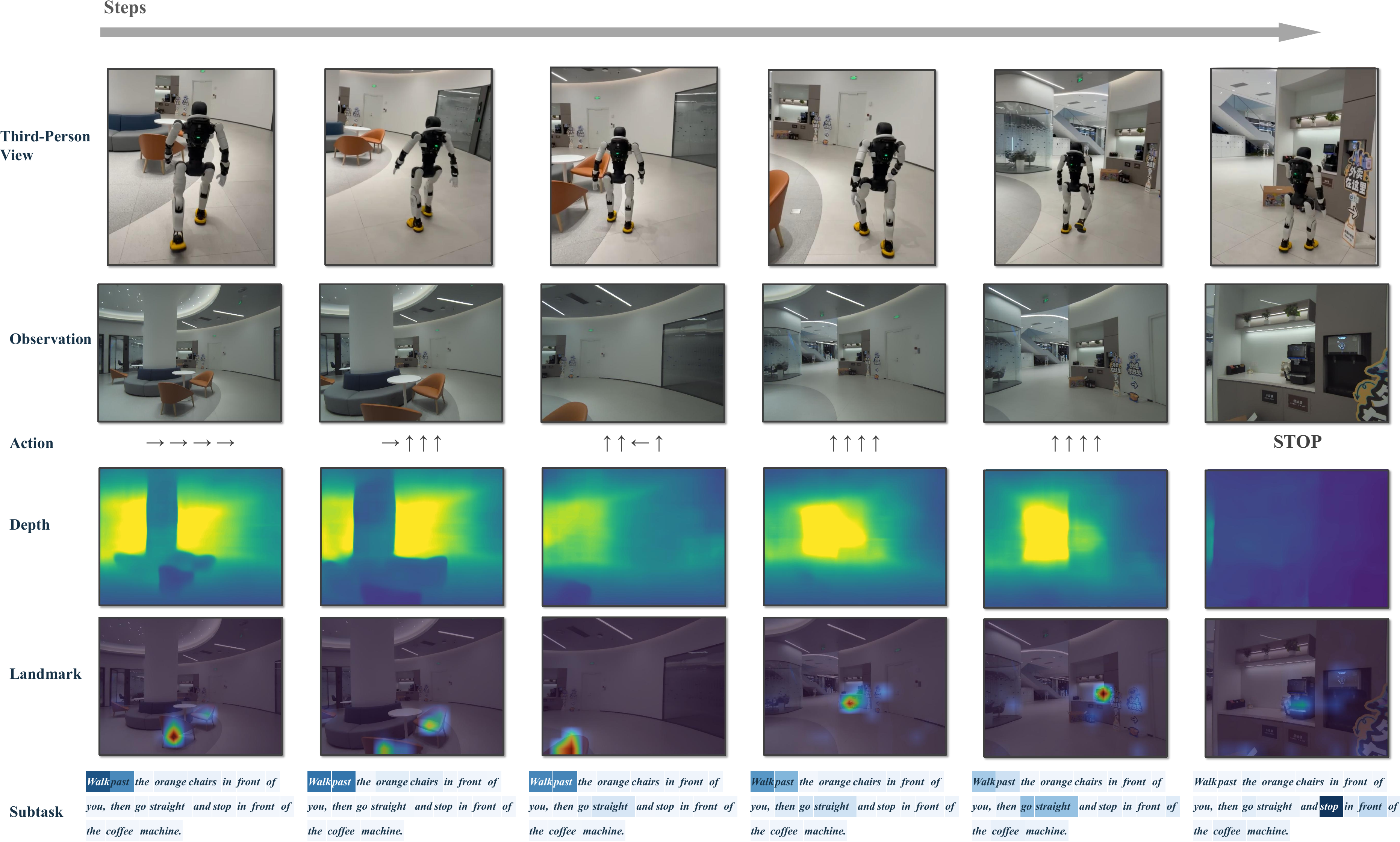}
\caption{\textbf{Real-World Navigation without Real-World Fine-Tuning.}
Six decisions show third-person views, RGB observations, actions, decoded depth, and pre-action attention. Depth is for analysis only; attention is normalized within each displayed token group.}
\label{fig:real_world}
\end{figure}

On an AgiBot X2 humanoid with a forward-facing RGB camera, GuidedNav navigates without real-world fine-tuning.
Figure~\ref{fig:real_world} shows the robot passing the orange chairs and stopping at the coffee machine: attention shifts from the intermediate landmark to the destination as the instruction progresses.
This qualitative transfer illustrates the learned instruction--landmark connection beyond simulation, while preserving direct action prediction.
Deployment details and additional cases appear in Appendix~\ref{app:real_rollouts}.

%% file: tables/main_results.tex
\begin{table}[t]
\centering
\caption{\textbf{Monocular RGB Navigation on R2R-CE and RxR-CE Val-Unseen.}
External data counts samples beyond standard R2R/RxR-CE data.}
\label{tab:main_results}
\setlength{\tabcolsep}{3pt}
\renewcommand{\arraystretch}{1.12}
\resizebox{\textwidth}{!}{%
\begin{tabular}{lrrrrrrrrc}
\toprule
\multirow{2}{*}{Method}
& \multicolumn{4}{c}{R2R Val-Unseen}
& \multicolumn{4}{c}{RxR Val-Unseen}
& \multirow{2}{*}{\shortstack{External Training\\Data}}\\
\cmidrule(lr){2-5}\cmidrule(lr){6-9}
& NE$\downarrow$ & OSR$\uparrow$ & SR$\uparrow$ & SPL$\uparrow$
& NE$\downarrow$ & SR$\uparrow$ & SPL$\uparrow$ & nDTW$\uparrow$ &\\
\midrule
NaVid\textsubscript{\scriptsize [RSS'24]}~\citep{zhang2024navid}
& 5.47 & 49.1 & 37.4 & 35.9
& -- & -- & -- & -- & 953K \\

Sim2Real\textsubscript{\scriptsize [CoRL'24]}~\citep{wang2025sim2real}
& 5.95 & 55.8 & 44.9 & 30.4
& 8.79 & 36.7 & 25.5 & 18.1 & 0K \\

Uni-NaVid\textsubscript{\scriptsize [RSS'25]}~\citep{zhang2025uninavid}
& 5.58 & 53.3 & 47.0 & 42.7
& 6.24 & 48.7 & 40.9 & -- & 3577K \\

InternVLA-N1\textsubscript{\scriptsize [Tech. Rep.'25]}~\citep{internrobotics2025internvlan1}
& 4.89 & 60.6 & 55.4 & 52.1
& 6.41 & 49.5 & 41.8 & 62.6 & 51200K \\

StreamVLN$^{*}$\textsubscript{\scriptsize [ICRA'26]}~\citep{wei2026streamvln}
& 6.05 & 53.8 & 45.5 & 41.6
& -- & -- & -- & -- & 10033K \\

NaVILA$^{*}$\textsubscript{\scriptsize [RSS'25]}~\citep{cheng2025navila}
& 5.37 & 57.6 & 49.7 & 45.5
& -- & -- & -- & -- & 12574K \\

NaVILA\textsubscript{\scriptsize [RSS'25]}~\citep{cheng2025navila}
& 5.22 & 62.5 & 54.0 & 49.0
& 6.77 & 49.3 & 44.0 & 58.8 & 13132K \\

Aux-Think\textsubscript{\scriptsize [NeurIPS'25]}~\citep{wang2025auxthink}
& 6.08 & 60.0 & 54.8 & 46.9
& 6.24 & 52.2 & 40.2 & -- & 1600K$^{\ddagger}$ \\

MonoDream\textsubscript{\scriptsize [AAAI'26]}~\citep{wang2026monodream}
& 5.45 & 61.5 & 55.8 & 49.1
& 6.38 & 49.4 & 40.9 & -- & 0K$^{\ddagger}$ \\

JanusVLN$^{\dagger}$\textsubscript{\scriptsize [ICLR'26]}~\citep{zeng2026janusvln}
& 5.17 & 58.0 & 52.8 & 49.2
& 6.46 & 51.4 & 44.3 & 59.1 & 0K \\

StreamVLN\textsubscript{\scriptsize [ICRA'26]}~\citep{wei2026streamvln}
& 4.98 & 64.2 & 56.9 & 51.9
& 6.22 & 52.9 & 46.0 & 61.9 & $\sim$26330K \\

NavFoM\textsubscript{\scriptsize [ICLR'26]}~\citep{zhang2026navfom}
& 5.01 & 64.9 & 56.2 & 51.2
& 5.51 & 57.4 & 49.4 & 60.2 & $\sim$10480K \\

JanusVLN\textsubscript{\scriptsize [ICLR'26]}~\citep{zeng2026janusvln}
& 4.78 & 65.2 & 60.5 & 56.8
& 6.06 & 56.2 & 47.5 & 62.1 & 10692K \\

\midrule
\rowcolor{blue!7}
\textbf{GuidedNav (Ours)}
& \textbf{3.83} & \textbf{71.5} & \textbf{65.4} & \textbf{59.1}
& \textbf{4.40} & \textbf{64.9} & \textbf{54.3} & \textbf{69.2}
& 3765K \\

\bottomrule
\end{tabular}%
}
\vspace{2pt}

\begin{minipage}{\textwidth}
\scriptsize
$\dagger$: VLN-CE expert trajectories only, without additional training data.
StreamVLN$^{*}$ uses EnvDrop data; NaVILA$^{*}$ excludes human-following data.
$\ddagger$: source-reported data counts use different definitions:
Aux-Think includes additional RxR, DAgger, and web data,
whereas MonoDream excludes simulator-generated data.
``--'' denotes an unavailable metric or data count, not zero.
\end{minipage}
\end{table}

%% file: tables/component_ablations.tex
\begin{table}[t]
\centering
\caption{Ablation studies on R2R-CE val-unseen.
SG denotes Spatial Guidance.
The upper block evaluates individual modules against Base Model, defined in Section~\ref{sec:setup};
the lower block progressively builds Spatial Guidance.
The complete configuration in each block is shaded in blue.}
\label{tab:ablations}
\small
\setlength{\tabcolsep}{5pt}
\renewcommand{\arraystretch}{1.15}

\begin{tabularx}{\linewidth}{@{}>{\raggedright\arraybackslash}X|rrrr@{}}
\toprule
Configuration
& NE$\downarrow$
& OSR$\uparrow$
& SR$\uparrow$
& SPL$\uparrow$ \\
\midrule

Base Model
& 4.87 & 63.8 & 59.3 & 56.1 \\

\midrule
\textit{Overall Guidance Ablation}
& & & & \\
\addlinespace[3pt]

Base + Subtask Attention Guidance
& 4.34 & \textbf{72.3} & 62.9 & 55.2 \\
Base + Landmark Attention Guidance
& 4.38 & 72.1 & 62.6 & 55.8 \\
Base + Spatial Guidance
& 4.33 & 68.5 & 62.4 & 57.2 \\

\rowcolor{blue!7}
\textbf{GuidedNav (All)}
& \textbf{3.83} & 71.5 & \textbf{65.4} & \textbf{59.1} \\

\midrule
\textit{Detailed Spatial Guidance Ablation}
& & & & \\
\addlinespace[3pt]

Base + Long-Horizon SG
& 4.35 & 68.4 & 61.8 & 56.6 \\
Base + Long-Horizon SG + Gated Fusion
& \textbf{4.30} & 68.0 & 62.0 & 56.7 \\

\rowcolor{blue!7}
\textbf{Base + Long-Horizon SG + Gated Fusion + Local SG}
& 4.33 & \textbf{68.5} & \textbf{62.4} & \textbf{57.2} \\

\bottomrule
\end{tabularx}
\end{table}

%% file: tables/attn_vis.tex
\begin{table}[t]
\centering
\caption{\textbf{Attention Localization and Spatial Probing.}
Subtask and landmark localization are evaluated on R2R-CE val-unseen;
EBPG is computed at the patch level, and Rank Acc.\ measures near--far depth ordering.
Protocols appear in Appendices~\ref{app:attention_localization}
and~\ref{app:spatial_probes}.}
\label{tab:attention_localization}
\label{tab:spatial_probe}
\small
\setlength{\tabcolsep}{4pt}
\renewcommand{\arraystretch}{1.15}
\resizebox{\linewidth}{!}{%
\begin{tabular}{@{}lcc|cc|ccc@{}}
\toprule
& \multicolumn{2}{c|}{\textit{Subtask Localization}}
& \multicolumn{2}{c|}{\textit{Landmark Localization}}
& \multicolumn{3}{c}{\textit{Spatial Probing}} \\
\cmidrule(lr){2-3}
\cmidrule(lr){4-5}
\cmidrule(lr){6-8}
Model
& AUPRC (\%) $\uparrow$ & AUROC (\%) $\uparrow$
& PG (\%) $\uparrow$ & EBPG (\%) $\uparrow$
& AbsRel $\downarrow$ & RMSE (m) $\downarrow$
& Rank Acc.\ (\%) $\uparrow$ \\
\midrule
Base Model
& 34.5 & 53.2 & 15.2 & 13.4
& 0.25 & 0.82 & 85.9 \\
GuidedNav
& \textbf{56.3} & \textbf{76.1}
& \textbf{49.1} & \textbf{42.6}
& \textbf{0.19} & \textbf{0.64} & \textbf{91.7} \\
\bottomrule
\end{tabular}%
}
\end{table}

%% file: conclusion.tex
\section{Conclusion}
\label{sec:conclusion}
We presented GuidedNav, which unifies instruction progress, landmark grounding, and spatial geometry through training-time representation guidance.
Supervising native pre-action attention and transferring geometry across long-horizon and local contexts strengthen the policy while preserving direct action prediction.
Guided-R2R and Guided-RxR supply aligned subtask and landmark supervision.
State-of-the-art monocular RGB results, representation analyses, and real-world transfer demonstrate the promise of this approach: richer navigation representations without an online geometric teacher or intermediate textual reasoning.

%% file: appendix.tex

\setlength{\textfloatsep}{8pt plus 2pt minus 2pt}
\setlength{\floatsep}{7pt plus 2pt minus 2pt}
\setlength{\intextsep}{7pt plus 2pt minus 2pt}
\setlength{\parskip}{4pt plus 1pt minus 1pt}
\captionsetup{font=small,skip=4pt}
\raggedbottom
\section{Training and Implementation Details}
\label{app:implementation}
\label{app:spatial_architecture}

\subsection{Three-Stage Training and Baseline Definition}
\label{app:training_recipe}
Training starts from Qwen3-VL-8B-Instruct and follows Table~\ref{tab:training_stages}.
Stages 1 and 2 use expert and DAgger navigation data, respectively, each mixed with approximately 0.79M VQA examples from LLaVA-Video-178K and VSI-590K.
Guidance is introduced only in Stage 3, using the original R2R and English RxR expert data enriched with our annotations (Appendix~\ref{app:data_engine}); this stage excludes DAgger and VQA data.

\begin{table}[H]
\centering\small
\caption{\textbf{Three-Stage Training Recipe.} Every stage uses one epoch and global batch size 256. Stages 1 and 2 use full fine-tuning; Stage 3 adapts the policy with rank-32 LoRA and scaling factor 16.}
\label{tab:training_stages}
\setlength{\tabcolsep}{4pt}
\renewcommand{\arraystretch}{1.12}
\begin{tabularx}{\linewidth}{@{}lXXX@{}}
\toprule
 & Stage 1 & Stage 2 & Stage 3 (GuidedNav)\\
\midrule
Initialization & Qwen3-VL-8B-Instruct & Stage 1 checkpoint & Stage 2 checkpoint\\
Navigation data & R2R + RxR experts & DAgger & Annotated R2R + RxR experts\\
VQA data & Included & Included & Not included\\
Policy update & Full parameters & Full parameters & LoRA adapters\\
Peak policy LR & $2\times10^{-5}$ & $1\times10^{-5}$ & $2\times10^{-4}$\\
Our guidance & None & None & Subtask, landmark, spatial\\
\bottomrule
\end{tabularx}
\end{table}

\paragraph{Optimization.}
All stages use AdamW, cosine decay, 3\% learning-rate warmup, weight decay 0.01, BF16, and DeepSpeed ZeRO-2.
Stages 1 and 2 use Adam coefficients $(0.9,0.999)$, maximum gradient norm 1.0, and a peak learning rate of $5\times10^{-6}$ for the multimodal projector and vision tower.
Their maximum sequence length is 8,192 tokens, with image pixel bounds of $28^2\times16$ and $28^2\times256$.
The random seed is 42.

\paragraph{Base Model and component comparisons.}
Base Model applies one epoch of Stage-3 LoRA fine-tuning to the Stage-2 model on the same R2R and RxR expert data, without the proposed guidance modules.
It shares GuidedNav's adapter rank, scaling factor, learning rate, global batch size, and epoch budget.
This baseline is used for component ablations, route-length analysis, and representation comparisons.
Component variants add the indicated modules; GuidedNav combines all three.

\subsection{Spatial Architecture and Auxiliary Settings}

\begin{figure}[t]
\centering
\includegraphics[width=0.85\linewidth]{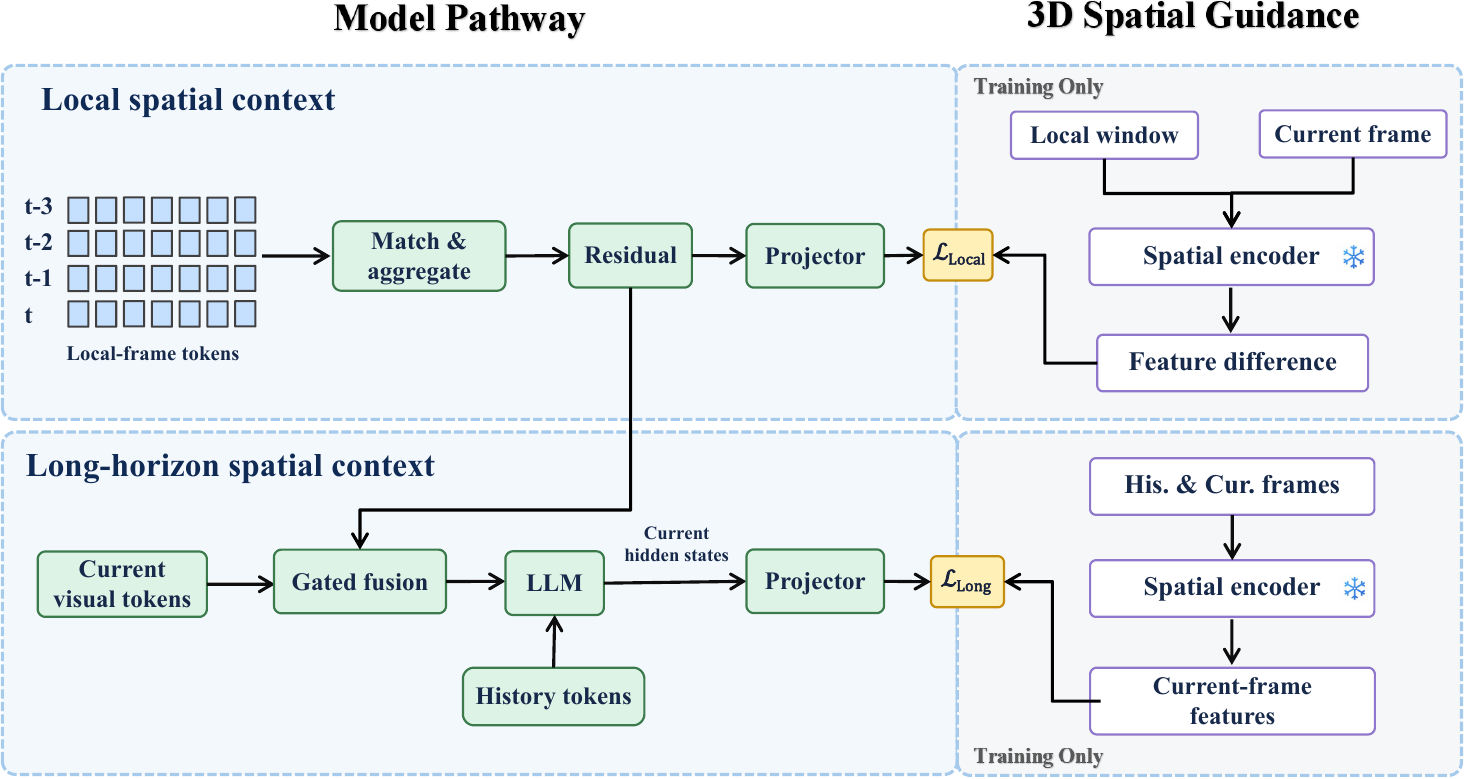}
\caption{\textbf{Long-Horizon Spatial Guidance and Local Spatial Refinement.}
The upper branch supervises matched-token residuals with $\mathcal{L}_{\mathrm{Local}}$ and fuses them into current visual tokens.
The lower branch aligns current hidden states with history-conditioned teacher features through $\mathcal{L}_{\mathrm{Long}}$.
The teacher and auxiliary projectors are used only during training; temporal fusion remains active at inference.}
\label{fig:spatial}
\end{figure}

Figure~\ref{fig:spatial} summarizes long-horizon alignment and local refinement.
The long-horizon teacher receives the current frame and up to seven sampled past observations; alignment covers the full current patch grid.
The local branch uses the current frame and up to three preceding observations.
Both branches use only observations available at time $t$.

\paragraph{Auxiliary supervision.}
Attention is read from decoder block 27 (one-based), heads 0 and 16 (zero-based), and normalized separately over instruction tokens and current-image patches.
Spatial supervision uses the output hidden states of the same block.
The weights $(\lambda_{\mathrm{sub}},\lambda_{\mathrm{lm}},\lambda_{\mathrm{long}},\lambda_{\mathrm{local}})$ are $(0.01,0.01,0.5,0.1)$, with 3\% auxiliary warmup.
Missing annotations disable only the corresponding losses.

\paragraph{Teacher features and grid alignment.}
\label{app:feature_alignment}
Both branches use frozen VGGT-1B, concatenating frame and global features from its final aggregator output and removing camera and register tokens to obtain 2048-dimensional patch features.
The teacher's 14-pixel patch grid is reshaped spatially and bilinearly resized to the policy's post-merger visual-token grid, obtained from its image-grid metadata and $2\times2$ merge factor.
The long-horizon target adds VGGT's UV positional embedding with scale 0.1 before resizing; local and single-frame targets omit this embedding.
Separate two-layer MLPs with GELU map the 4096-dimensional student features to 2048 dimensions.

\paragraph{Local geometric target.}
Features from the local window and current frame alone are independently normalized before subtraction, and the difference is normalized again:
\begin{equation}
    \Delta f_{t,j} = \mathcal{N}\!\left(
    \mathcal{N}(f_{t,j}^{\mathrm{local}})
    -\mathcal{N}(f_{t,j}^{\mathrm{single}})\right),
    \label{eq:local_target}
\end{equation}
where $\mathcal{N}$ denotes $\ell_2$ normalization.

\paragraph{Sparse temporal matching.}
\label{app:fusion_matching}
For a past-frame token $i$ and current token $j$, matching uses detached features and the score
\begin{equation}
 s_{u,i,j}=\cos(z_{u,i},z_{t,j})-0.05\|q_i-q_j\|_2^2,
\end{equation}
where $q_i,q_j$ are normalized coordinates on the resized token grid.
A pair is accepted only if each token is the other's highest-scoring match.
We retain at most $\operatorname{round}(0.25N_t)$ highest-scoring mutual pairs per past frame, with weights $0.2,0.35,0.5$ for $t-3,t-2,t-1$, omitting unavailable frames.
Gradients pass through selected features but not discrete matching.
The scalar gate is $g=0.2\operatorname{sigmoid}(a)$, initialized to $0.05$.
The local loss supervises the residual before gating; action and other downstream objectives train the gate through the fused tokens.

\paragraph{Subtask interval balancing.}
\label{app:supervision_weights}
For each source dataset, let $V$ count valid span-labeled examples, $G$ count trajectory--subtask groups, and $n_g$ count valid examples in group $g$.
Equation~\ref{eq:subtask_loss} uses $c_t=V/(Gn_g)$, giving each interval equal total weight and each valid example unit weight on average.
Landmark supervision is averaged over valid frame-level masks without interval balancing.

\begin{table}[H]
\centering\small
\caption{\textbf{Simulator and Action Settings.} The common action settings apply to both datasets; collision-body dimensions shown below are the RxR-specific settings.}
\label{tab:simulator_settings}
\begin{tabular}{lr}
\toprule
Setting & Value\\
\midrule
Decoding & Greedy\\
Maximum sampled context images & 8\\
Executed action chunk & Up to 4 primitives\\
Forward displacement & 0.25 m\\
Turn angle & $15^\circ$\\
Maximum episode actions & 500\\
Sliding & Enabled\\
RGB camera height & 1.25 m\\
RxR collision-body height / radius & 0.88 m / 0.18 m\\
\bottomrule
\end{tabular}
\end{table}

\clearpage
\section{Landmark-Centric Data Engine}
\label{app:data_engine}
Guided-R2R and Guided-RxR augment R2R and RxR~\citep{anderson2018vln,ku2020rxr} with active subtask spans and landmark masks (Figure~\ref{fig:annotation_pipeline}).
R2R reuses FGR2R subtasks~\citep{hong2020subinstruction}; gpt-oss-120b decomposes English RxR training instructions into spans, landmarks, and completion conditions.
RxR word timestamps and guide poses provide coarse temporal alignment, which GLM-4.5V refines using trajectory overviews and frames near subtask boundaries.
For each aligned clip, GLM-4.5V selects explicitly mentioned landmarks or visually supported structures implied by the action, verifies GroundingDINO proposals~\citep{liu2024groundingdino}, and selects anchor frames.
SAM2~\citep{ravi2025sam2} propagates accepted box prompts within each interval.
Masks are filtered by area, detection consistency, and temporal support; the original observations, instructions, and action targets are preserved.

\paragraph{Mapping R2R subtasks to continuous trajectories.}
We match exported episodes to FGR2R instructions and reference paths.
Reference-path anchors are associated with their closest recorded continuous poses using Euclidean distance.
Anchor-arrival indices are made nondecreasing when necessary; the FGR2R subpath boundaries then define the per-frame subtask assignments.
Exports with missing poses, ambiguous matches, or incompatible path topology are excluded from this alignment.
\paragraph{Annotation access.}
Offline annotation can inspect an entire expert trajectory to establish temporal intervals.
The policy and spatial teacher use only current and past observations at each decision.
Samples without reliable spans or masks contribute to action supervision, with the unavailable auxiliary terms omitted.

\subsection{Guided-R2R and Guided-RxR Examples}
\label{app:dataset_examples}
\begin{figure}[H]
\centering
\includegraphics[width=\linewidth]{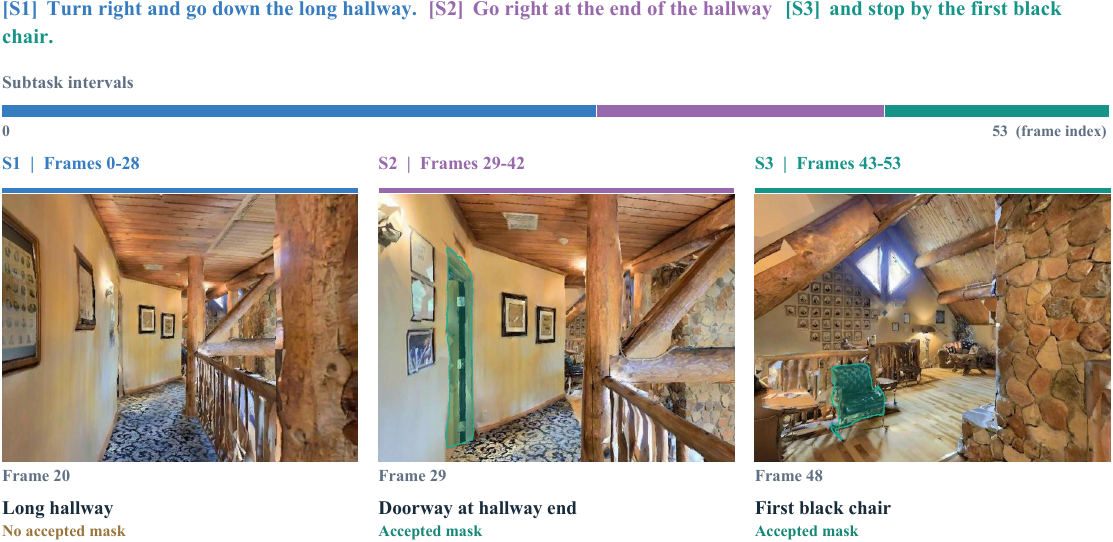}
\caption{\textbf{Guided-R2R Annotation Example.} Three subtask spans align hallway traversal, a right turn, and stopping by a black chair with their frame intervals. Representative observations show the selected landmarks and whether a mask was accepted.}
\label{fig:guided_r2r}
\end{figure}

\begin{figure}[H]
\centering
\includegraphics[width=\linewidth]{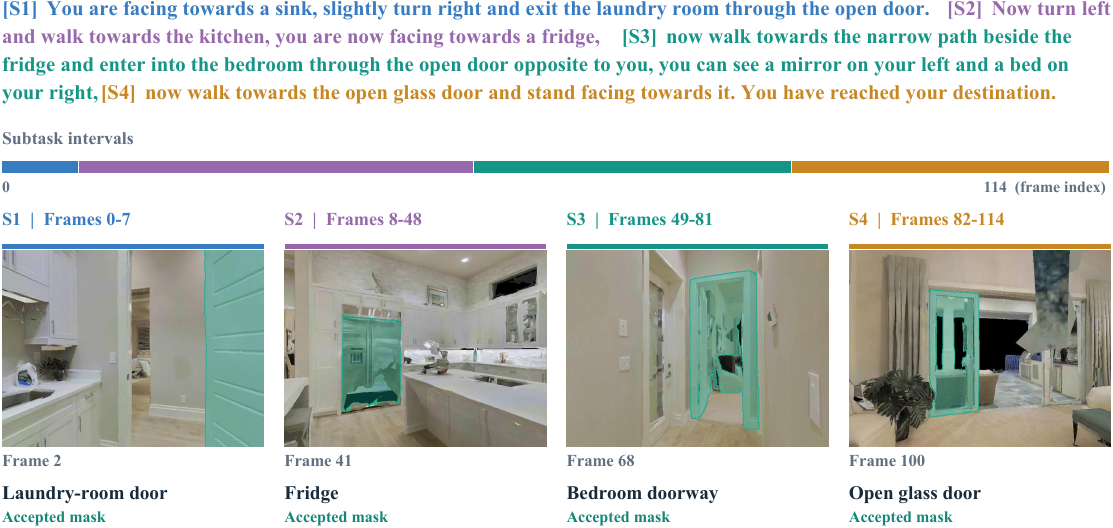}
\caption{\textbf{Guided-RxR Annotation Example.} Four subtask spans follow a route from a laundry room through a kitchen and bedroom to a glass door. Matching colors link instruction text, frame intervals, and representative landmark masks.}
\label{fig:guided_rxr}
\end{figure}

Figures~\ref{fig:guided_r2r} and~\ref{fig:guided_rxr} link subtask spans, frame intervals, and landmark masks by color.
The R2R hallway interval illustrates missing mask coverage: action supervision is retained, while landmark supervision is omitted.

\clearpage
\section{Further Analysis of the Guidance Modules}
\label{app:module_analysis}

\subsection{Subtask Attention Guidance}
\label{app:subtask_analysis}
Table~\ref{tab:subtask_ablation} compares textual supervision with native attention guidance.
\textit{Sequential Generation} predicts the active subtask before the action sequence, using teacher forcing and cross-entropy weights of 0.1 and 1.0 for subtask and action tokens, respectively, normalized by total token weight.
\textit{Auxiliary Generation} uses separate action and subtask forward passes through the shared policy, with independently normalized losses and a text-loss weight of 0.1 warmed up at the start of training.
Its generated text is not fed to the action branch, and inference uses only the action pass.
\textit{Attention Guidance} supervises pre-action attention to the active span.
It gives the highest SR and lowest NE, while Auxiliary Generation gives the highest OSR and Base Model the highest SPL.
Figure~\ref{fig:subtask_attention} visualizes attention as instruction progress changes.

\input{tables/subtask}

\subsection{Landmark Attention Guidance}
\label{app:landmark_analysis}
\begin{figure}[t]
\centering
\includegraphics[width=\linewidth]{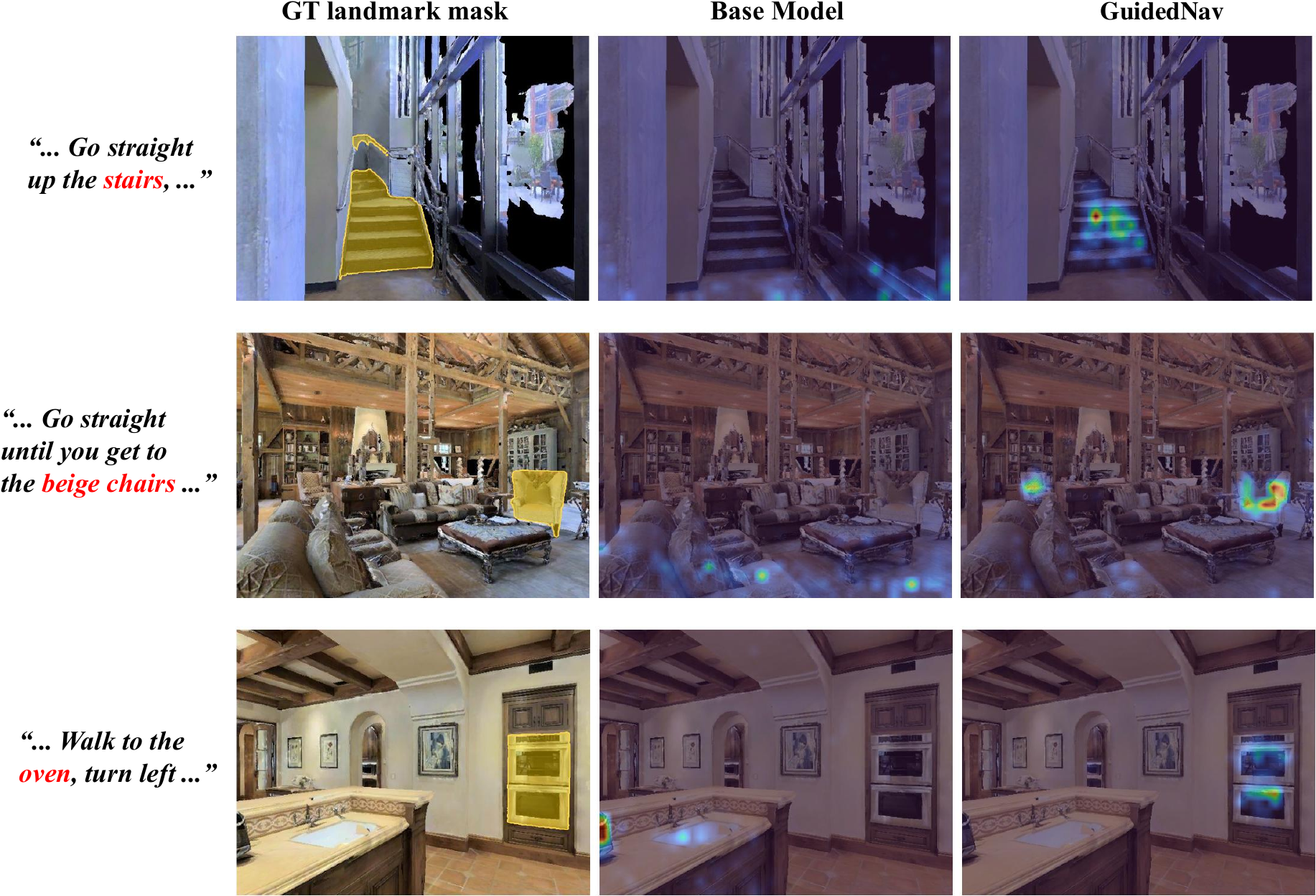}
\caption{\textbf{Landmark Attention Localization.}
Instruction excerpts and reference masks are compared with Base Model and GuidedNav pre-action attention for stairs, chairs, and an oven.
Heatmaps cover the full current image.}
\label{fig:landmark_attention}
\end{figure}

Table~\ref{tab:landmark_ablation} compares landmark supervision strategies.
Our objective increases attention mass within accepted landmark regions without fixing its distribution over patches.
It gives the highest SR and OSR and lowest NE, while Base Model retains the highest SPL.
Figure~\ref{fig:landmark_attention} illustrates localization relative to the reference masks.

\input{tables/landmark}

\subsection{Quantitative Attention Localization}
\label{app:attention_localization}
Table~\ref{tab:attention_localization} evaluates correspondence with automatically annotated instruction spans and landmarks.
For subtasks, active-span tokens are positives and attention weights provide scores for area under the precision--recall curve (AUPRC) and receiver operating characteristic curve (AUROC)~\citep{davis2006prroc}.
Pointing Game (PG)~\citep{zhang2016excitation} checks whether the peak-attention location lies inside a landmark, while patch-level Energy-Based Pointing Game (EBPG)~\citep{wang2020scorecam} measures the fraction of attention mass supported by the landmark mask.
All four metrics improve for GuidedNav (Table~\ref{tab:attention_localization}). Attention is normalized separately over instruction tokens and current-image patches, so each metric describes localization within its corresponding token group.

\FloatBarrier
\subsection{Spatial Representation Probing}
\label{app:spatial_probes}
We fit depth probes independently to frozen Base Model and GuidedNav representations; the probes are excluded from action prediction.
Table~\ref{tab:spatial_probe} reports absolute relative error (AbsRel) and root mean squared error (RMSE)~\citep{eigen2014depth}, together with near--far ranking accuracy, the fraction of correctly ordered depth pairs inspired by relative-depth evaluation~\citep{chen2016depthwild}.
Lower errors and higher ranking accuracy indicate that geometric information is more accessible under the fitted probes.

\paragraph{Dense depth visualization.}
For qualitative visualization, a separate convolutional head decodes frozen current-frame features from decoder block 27 into $512\times384$ depth maps.
The visualization decoder is trained for one epoch on 10,000 R2R training frames, with 1,000 held-out frames from disjoint scenes.
Figure~\ref{fig:depth_comparison} compares the decoded scene structure with reference depth.
The visualization decoder is an analysis tool and adds no cost to navigation inference.

\clearpage

\section{Additional Navigation Visualizations}
\label{app:rollouts}
\subsection{Simulation Rollouts}

\begin{figure}[H]
\centering
\includegraphics[width=\linewidth]{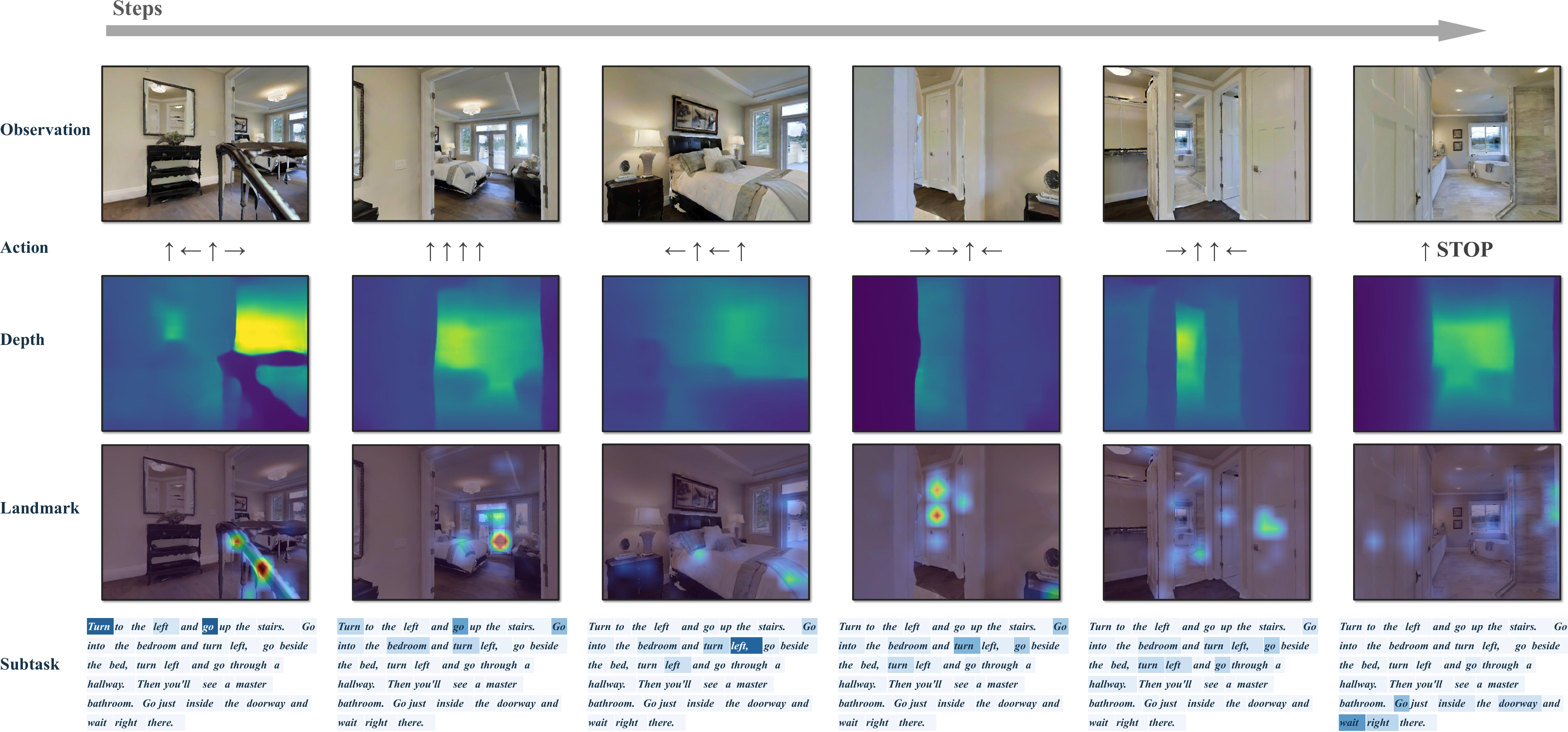}
\par\vspace{5pt}
\includegraphics[width=\linewidth]{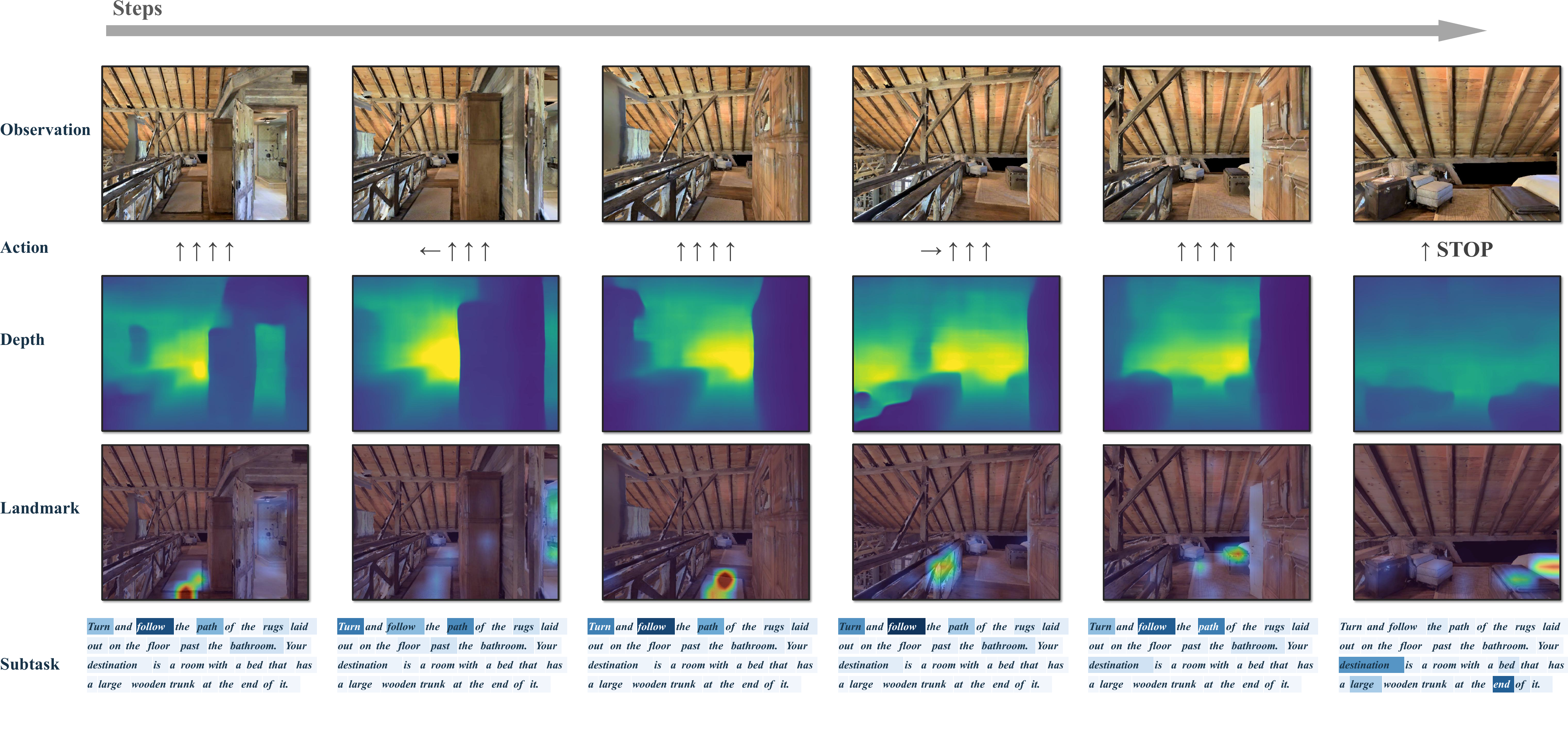}
\par\vspace{5pt}
\includegraphics[width=\linewidth]{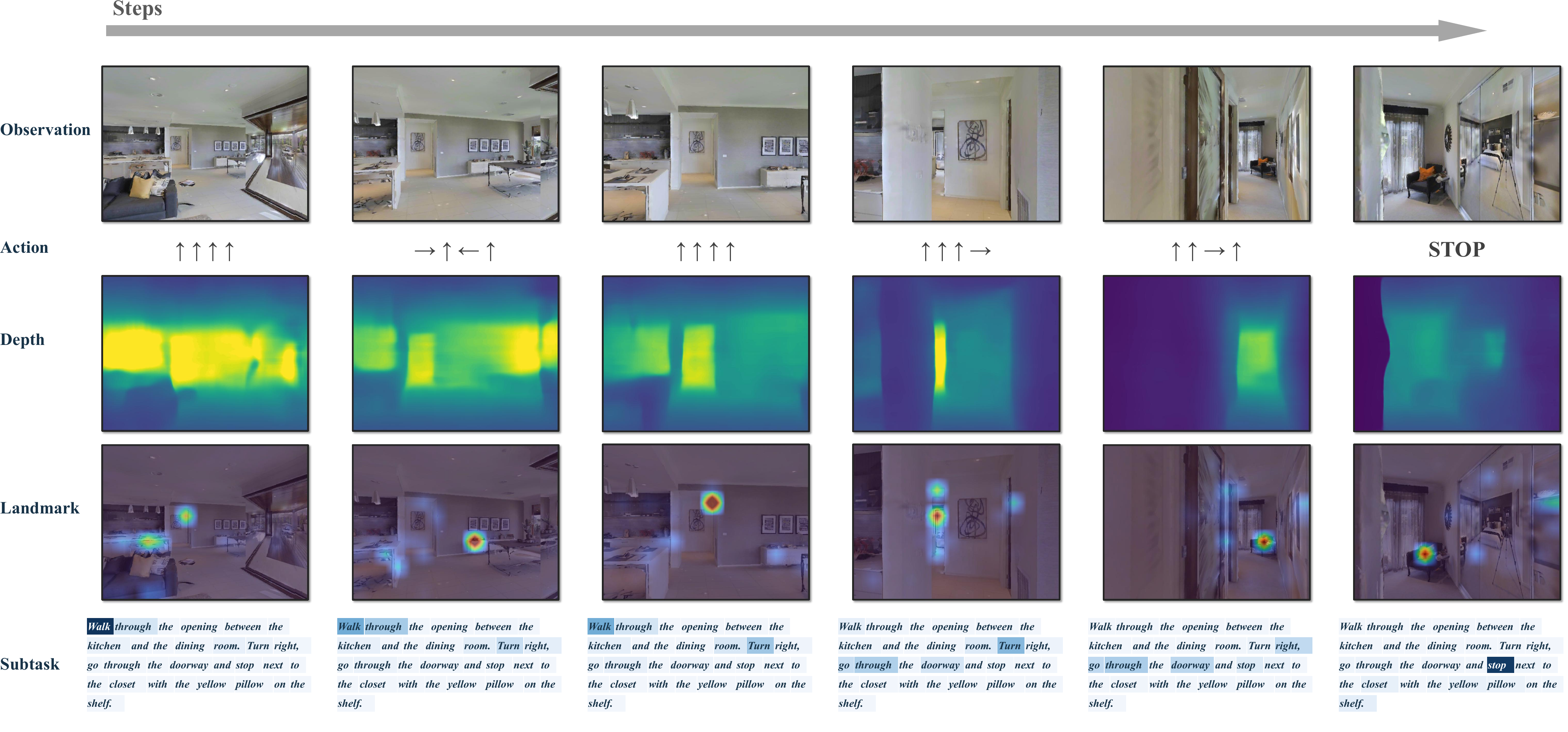}
\caption{\textbf{Successful R2R-CE Rollouts: Cases 1-3.}
Columns follow six sampled decisions along each trajectory. Rows show observations, predicted action chunks, decoded depth, landmark attention, and instruction attention; darker instruction highlighting indicates greater relative attention.}
\label{fig:sim_rollout_1}
\label{fig:sim_rollout_2}
\label{fig:sim_rollout_3}
\end{figure}
\clearpage

\begin{figure}[H]
\centering
\includegraphics[width=\linewidth]{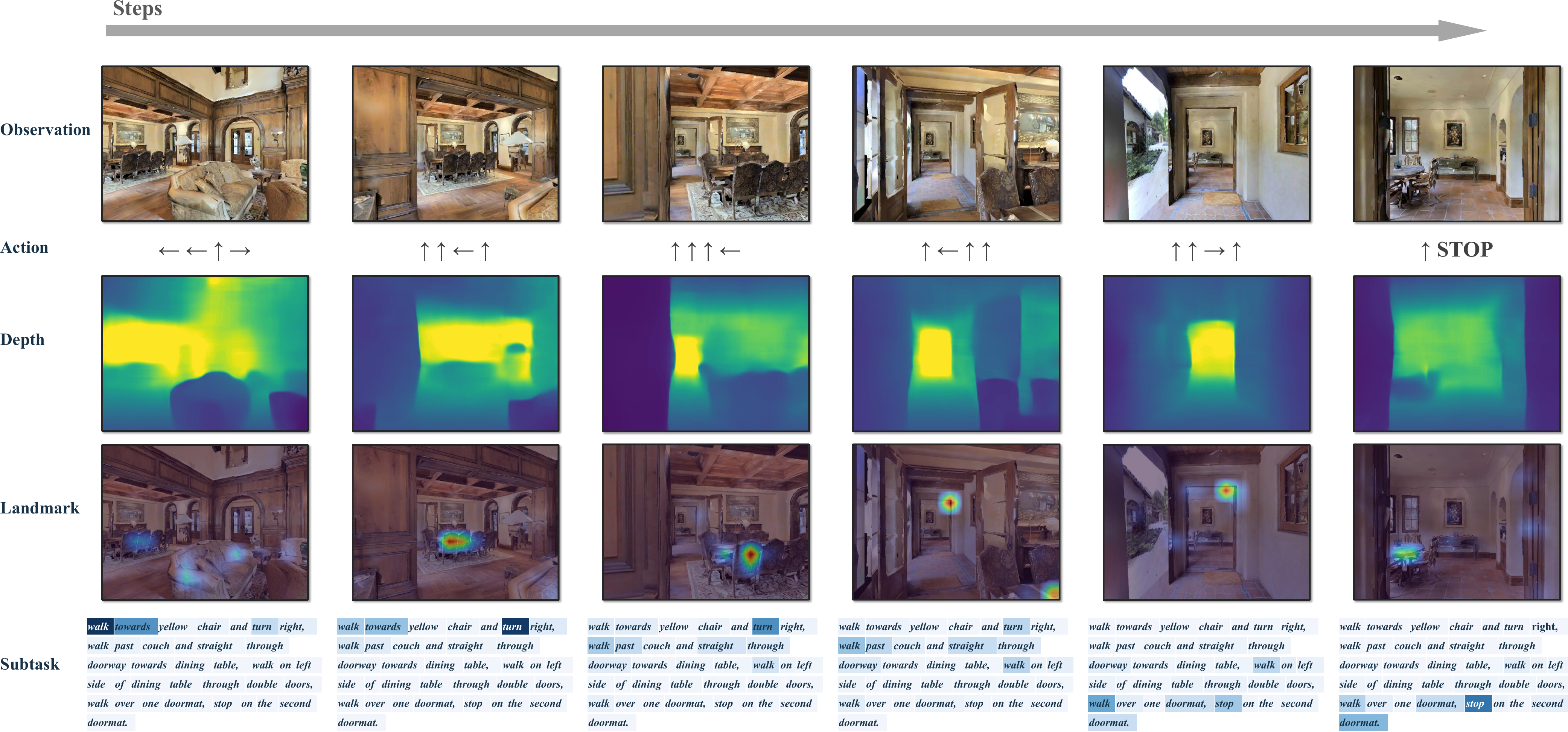}
\par\vspace{5pt}
\includegraphics[width=\linewidth]{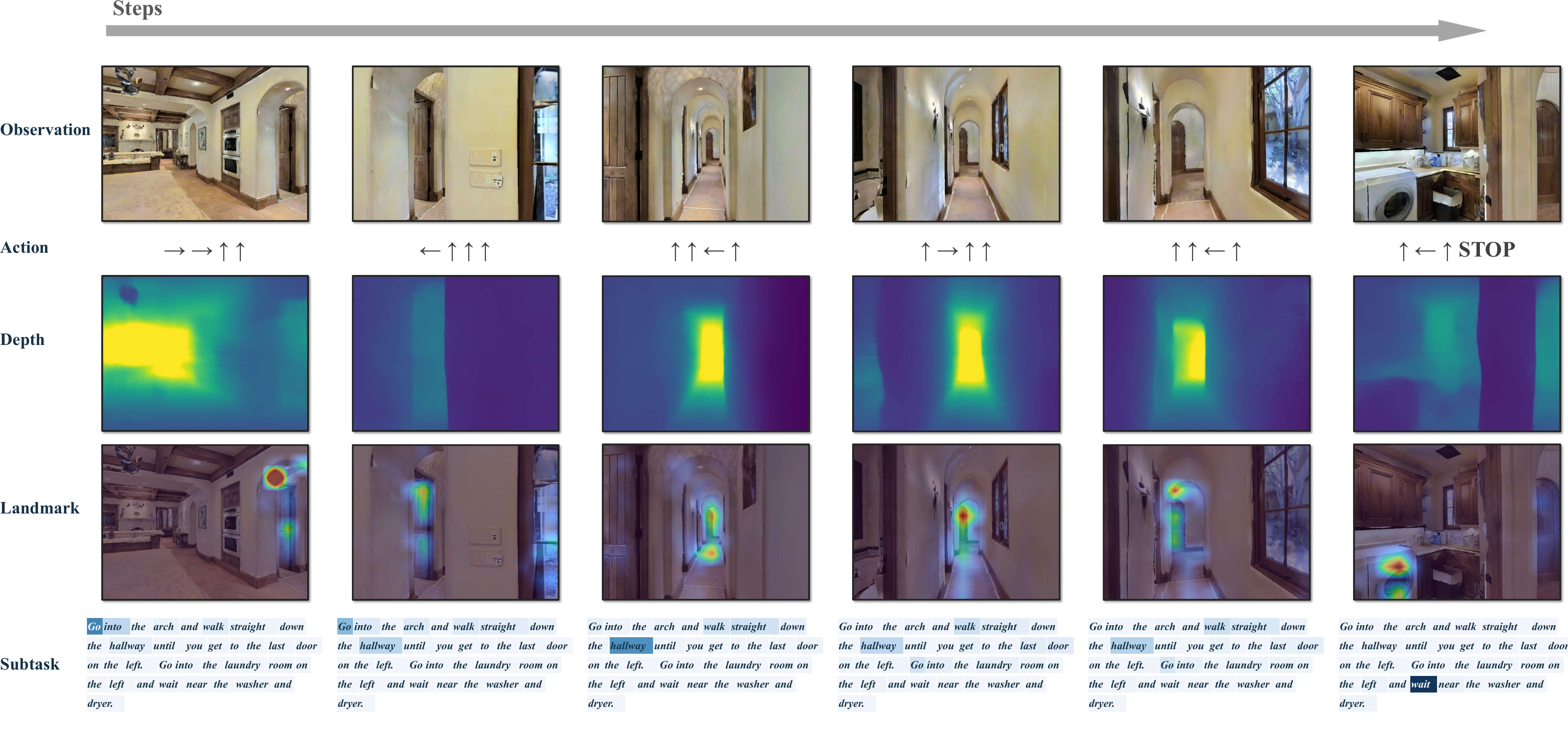}
\par\vspace{5pt}
\includegraphics[width=\linewidth]{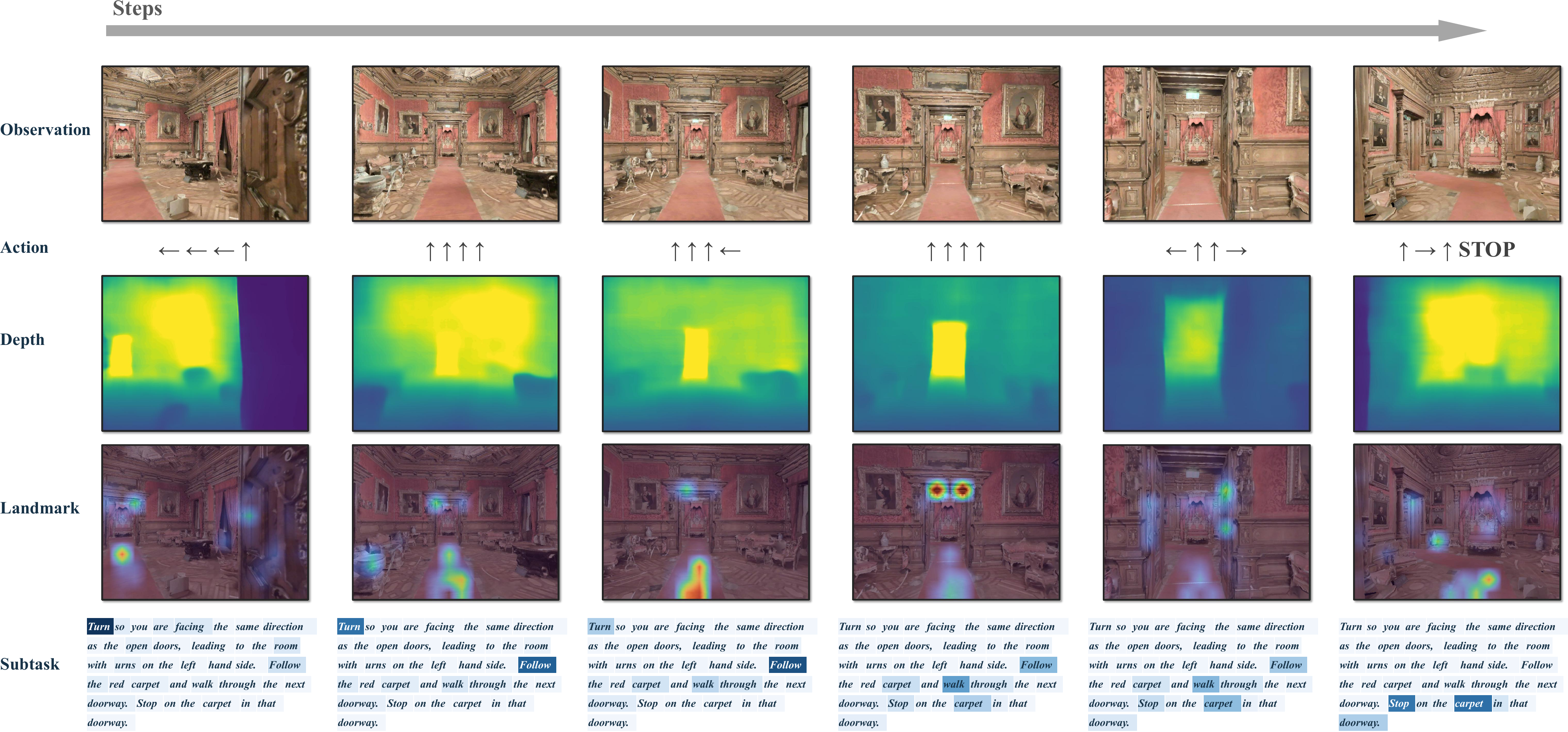}
\caption{\textbf{Successful R2R-CE Rollouts: Cases 4-6.}
Six sampled decisions per trajectory, following the conventions in Figure~\ref{fig:sim_rollout_1}.}
\label{fig:sim_rollout_4}
\label{fig:sim_rollout_5}
\label{fig:sim_rollout_6}
\end{figure}
\clearpage

\subsection{Real-World Setup and Rollouts}
\label{app:real_rollouts}
\begin{figure}[H]
\centering
\includegraphics[width=0.9\linewidth]{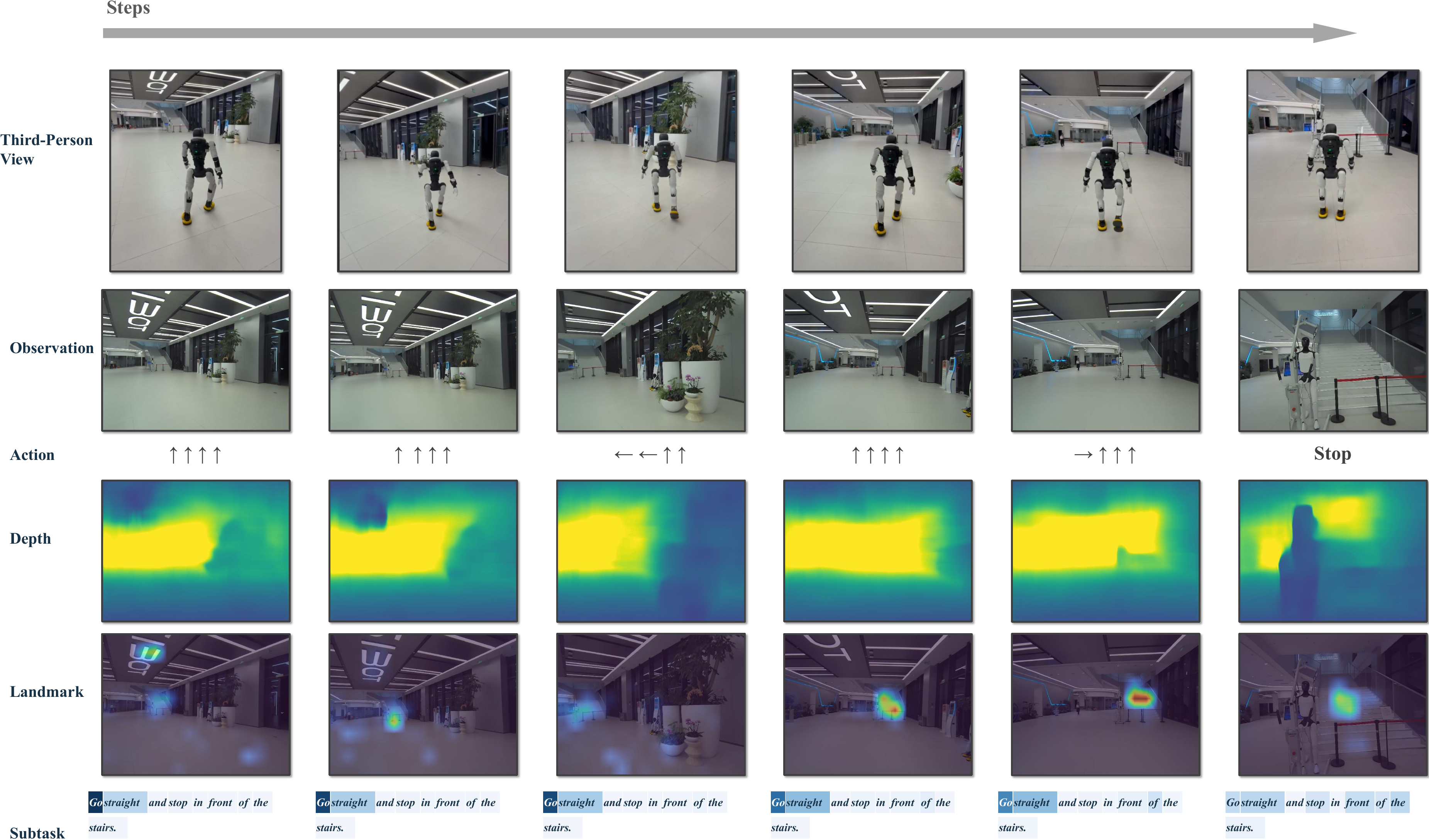}
\par\vspace{5pt}
\includegraphics[width=0.9\linewidth]{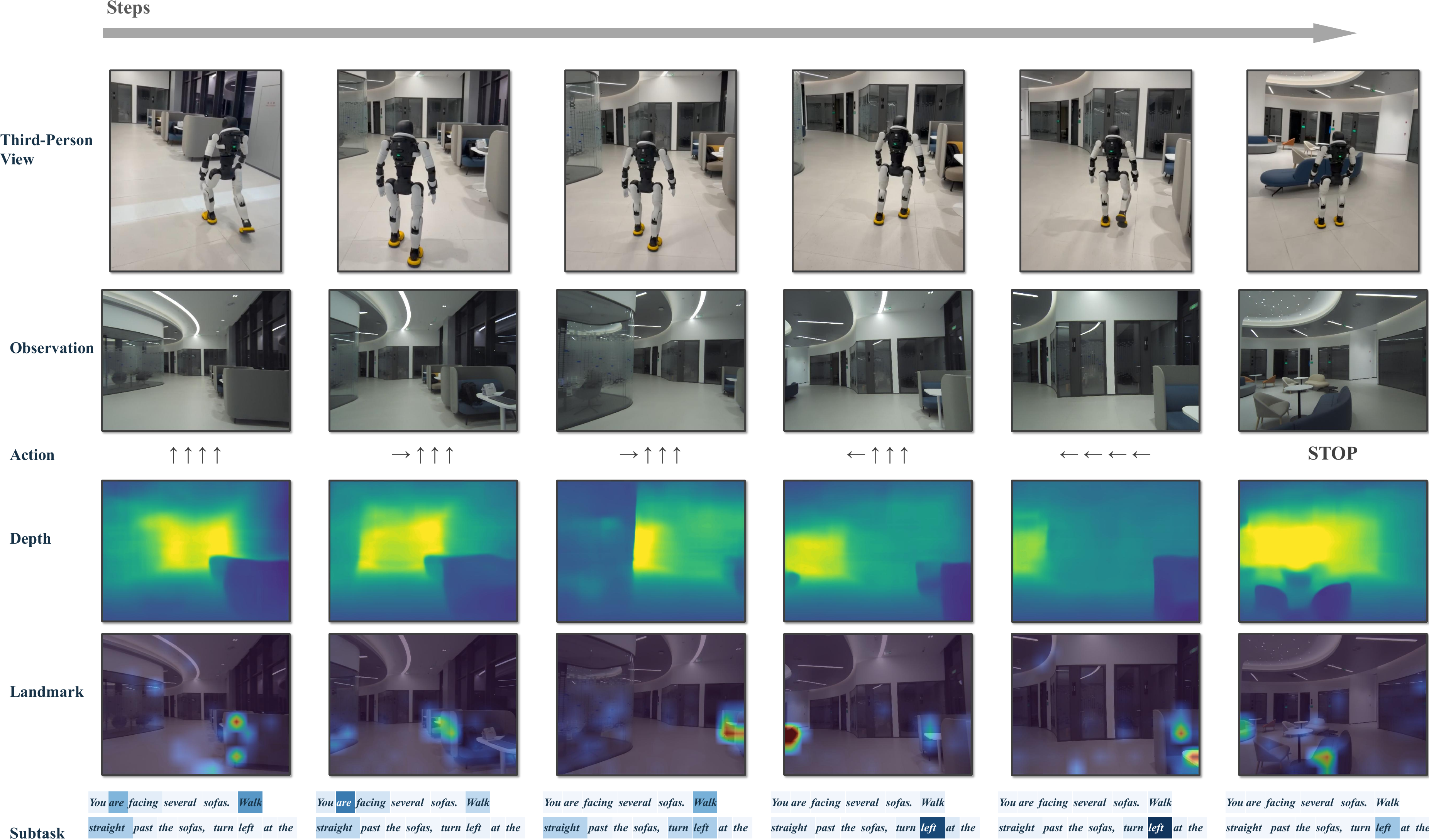}
\caption{\textbf{Additional Real-World Navigation Examples.}
Six sampled decisions per trajectory show observations, predicted actions, decoded depth, and native pre-action attention.
The visualization follows the conventions in Figure~\ref{fig:real_world}.}
\label{fig:real_rollout_stairs}
\label{fig:real_rollout_sofa}
\end{figure}

GuidedNav runs on an AgiBot X2 humanoid with a forward-facing RGB camera, receiving instructions and streaming egocentric observations.
An off-board NVIDIA RTX 4090 server transmits predicted actions to the robot; no real-world fine-tuning is used.
Depth is decoded only for visualization with the separate probe described in Appendix~\ref{app:spatial_probes}, and attention uses the readout in Appendix~\ref{app:spatial_architecture}.
Figure~\ref{fig:real_world} shows navigation past orange chairs to a coffee machine; Figure~\ref{fig:real_rollout_stairs} adds two trajectories in other scenes.
These examples provide qualitative evidence of transfer to real observations.

\clearpage
\section{Comparisons with Additional Observation Modalities}
\label{app:additional_results}
Table~\ref{tab:additional_results} compares methods using panoramic views or additional sensory inputs.
Their observation configurations differ from GuidedNav's monocular RGB input; monocular RGB results are reported in Table~\ref{tab:main_results}.
All baseline navigation results are taken from the respective papers.
\input{tables/additional_results}
\FloatBarrier

%% file: tables/subtask.tex
\begin{table}[H]
    \centering
    \caption{\textbf{Subtask Supervision Strategies on R2R-CE Val-Unseen.}}
    \label{tab:subtask_ablation}
    \small
    \begin{tabular}{lcccc}
        \toprule
        Method & SR $\uparrow$ & SPL $\uparrow$
        & OSR $\uparrow$ & NE $\downarrow$ \\
        \midrule
        Base Model
        & 59.3 & \textbf{56.1} & 63.8 & 4.87 \\
        Sequential Generation
        & 60.4 & 55.4 & 66.7 & 4.43\\
        Auxiliary Generation
        & 62.0 & 55.6 & \textbf{72.5} & 4.55 \\
        Attention Guidance
        & \textbf{62.9} & 55.2 & 72.3 & \textbf{4.34} \\
        \bottomrule
    \end{tabular}

\end{table}

%% file: tables/landmark.tex
\begin{table}[H]
    \centering
    \caption{\textbf{Landmark Attention Supervision on R2R-CE Val-Unseen.}
We compare Distribution Matching, Access Regularization, and our Attention Guidance objective. The latter maximizes attention mass supported by the landmark mask. SR, SPL, and OSR are percentages; NE is in meters. Bold denotes the best value in each column.}
    \label{tab:landmark_ablation}
    \small
    \begin{tabular}{lcccc}
        \toprule
        Method & SR $\uparrow$ & SPL $\uparrow$
        & OSR $\uparrow$ & NE $\downarrow$ \\
        \midrule
        Base Model
        & 59.3 & \textbf{56.1} & 63.8 & 4.87 \\
        Distribution Matching
        & 61.6 & 55.2 & 71.2 & 4.48 \\
        Access Regularization
        & 62.4 & 55.8 & 71.5 & 4.40 \\
        Attention Guidance (Ours)
        & \textbf{62.6} & 55.8 & \textbf{72.1} & \textbf{4.38} \\
        \bottomrule
    \end{tabular}
\end{table}

%% file: tables/additional_results.tex
\begin{table}[H]
\centering
\caption{\textbf{Additional Navigation Comparisons.} Methods using panoramic observations or additional sensory inputs. P: panoramic RGB; R: monocular RGB; O: odometry; D: depth. NE is in meters; all other metrics are percentages.}
\label{tab:additional_results}
\small
\setlength{\tabcolsep}{4pt}
\renewcommand{\arraystretch}{1.08}
\textbf{(a) R2R-CE val-unseen}\par\smallskip
\begin{tabularx}{\linewidth}{@{}>{\raggedright\arraybackslash}Xlrrrr@{}}
\toprule
Method & Inputs & NE$\downarrow$ & OSR$\uparrow$ & SR$\uparrow$ & SPL$\uparrow$\\
\midrule
HPN+DN~\citep{krantz2021waypoint} & P+O+D & 6.31 & 40.0 & 36.0 & 34.0 \\
CMA~\citep{hong2022bridging} & P+O+D & 6.20 & 52.0 & 41.0 & 36.0 \\
Sim2Sim~\citep{krantz2022sim2sim} & P+O+D & 6.07 & 52.0 & 43.0 & 36.0 \\
VLN$\circlearrowright$BERT~\citep{hong2022bridging} & P+O+D & 5.74 & 53.0 & 44.0 & 39.0 \\
Ego$^2$-Map~\citep{hong2023ego2map} & P+O+D & 5.54 & 56.0 & 47.0 & 41.0 \\
DreamWalker~\citep{wang2023dreamwalker} & P+O+D & 5.53 & 59.0 & 49.0 & 44.0 \\
GridMM~\citep{wang2023gridmm} & P+O+D & 5.11 & 61.0 & 49.0 & 41.0 \\
Reborn~\citep{an2022reborn} & P+O+D & 5.40 & 57.0 & 50.0 & 46.0 \\
InstructNav~\citep{long2025instructnav} & P+O+D & 6.89 & -- & 31.0 & 24.0 \\
\midrule
COSMO~\citep{zhang2025cosmo} & P & -- & 56.0 & 47.0 & 40.0 \\
AO-Planner~\citep{chen2025aoplanner} & P+D & 5.55 & 59.0 & 47.0 & 33.0 \\
LAW~\citep{raychaudhuri2021law} & O+D+R & 6.83 & 44.0 & 35.0 & 31.0 \\
MapNav~\citep{zhang2025mapnav} & O+D+R & 4.93 & 53.0 & 39.7 & 37.2 \\
g3D-LF~\citep{wang2025g3dlf} & O+D+R & 5.70 & 59.5 & 47.2 & 34.6 \\
Seq2Seq~\citep{krantz2020vlnce} & D+R & 7.77 & 37.0 & 25.0 & 22.0 \\
NaVid-4D~\citep{liu2025navid4d} & D+R & 5.99 & 55.7 & 43.8 & 37.1 \\
NavMorph~\citep{yao2025navmorph} & D+R & 5.75 & 56.9 & 47.9 & 33.2 \\
GA-VLN~\citep{yang2026gavln} & O+D+R & 4.80 & 67.6 & 61.0 & 55.2 \\
\bottomrule
\end{tabularx}
\par\medskip
\textbf{(b) RxR-CE val-unseen}\par\smallskip
\begin{tabularx}{\linewidth}{@{}>{\raggedright\arraybackslash}Xlrrrr@{}}
\toprule
Method & Inputs & NE$\downarrow$ & SR$\uparrow$ & SPL$\uparrow$ & nDTW$\uparrow$\\
\midrule
CMA~\citep{hong2022bridging} & P+O+D & 8.76 & 26.5 & 22.1 & 47.0 \\
VLN$\circlearrowright$BERT~\citep{hong2022bridging} & P+O+D & 8.98 & 27.0 & 22.6 & 46.7 \\
Reborn~\citep{an2022reborn} & P+O+D & 5.98 & 48.6 & 42.0 & 63.3 \\
AO-Planner~\citep{chen2025aoplanner} & P+D & 7.06 & 43.3 & 30.5 & 50.1 \\
LAW~\citep{raychaudhuri2021law} & O+D+R & 10.90 & 8.0 & 8.0 & 38.0 \\
Seq2Seq~\citep{krantz2020vlnce} & D+R & 12.10 & 13.9 & 11.9 & 30.8 \\
NavMorph~\citep{yao2025navmorph} & D+R & 8.85 & 30.8 & 22.8 & 44.2 \\
GA-VLN~\citep{yang2026gavln} & O+D+R & 5.88 & 55.4 & 45.2 & -- \\
\bottomrule
\end{tabularx}
\par\smallskip
\begin{minipage}{\linewidth}\footnotesize
Results are from the respective papers. ``--'' denotes an unavailable metric, not zero; methods without reported RxR results are omitted from panel (b). Comparable external-training-data counts are unavailable for these entries; GA-VLN uses additional navigation datasets. Monocular RGB-only results appear in Table~\ref{tab:main_results}.
\end{minipage}
\end{table}

%% file: references.bib
@inproceedings{zeng2026janusvln,
  title = {{JanusVLN}: Decoupling Semantics and Spatiality with Dual Implicit Memory for Vision-Language Navigation},
  author = {Zeng, Shuang and Qi, Dekang and Chang, Xinyuan and Xiong, Feng and Xie, Shichao and Wu, Xiaolong and Liang, Shiyi and Xu, Mu and Wei, Xing and Guo, Ning},
  booktitle = {International Conference on Learning Representations},
  year = {2026},
  url = {https://arxiv.org/abs/2509.22548}
}

@inproceedings{ku2020rxr,
  title = {Room-Across-Room: Multilingual Vision-and-Language Navigation with Dense Spatiotemporal Grounding},
  author = {Ku, Alexander and Anderson, Peter and Patel, Roma and Ie, Eugene and Baldridge, Jason},
  year = {2020},
  booktitle = {EMNLP},
  url = {https://arxiv.org/abs/2010.07954}
}

@inproceedings{krantz2020vlnce,
  title = {Beyond the Nav-Graph: Vision-and-Language Navigation in Continuous Environments},
  author = {Krantz, Jacob and Wijmans, Erik and Majumdar, Arjun and Batra, Dhruv and Lee, Stefan},
  year = {2020},
  booktitle = {ECCV},
  url = {https://arxiv.org/abs/2004.02857}
}

@inproceedings{zhang2024navid,
  title = {{NaVid}: Video-based {VLM} Plans the Next Step for Vision-and-Language Navigation},
  author = {Zhang, Jiazhao and Wang, Kunyu and Xu, Rongtao and Zhou, Gengze and Hong, Yicong and Fang, Xiaomeng and Wu, Qi and Zhang, Zhizheng and Wang, He},
  year = {2024},
  booktitle = {Robotics: Science and Systems},
  url = {https://arxiv.org/abs/2402.15852}
}

@inproceedings{cheng2025navila,
  title = {{NaVILA}: Legged Robot Vision-Language-Action Model for Navigation},
  author = {Cheng, An-Chieh and Ji, Yandong and Yang, Zhaojing and Zou, Xueyan and Kautz, Jan and Biyik, Erdem and Yin, Hongxu and Liu, Sifei and Wang, Xiaolong},
  year = {2025},
  booktitle = {Robotics: Science and Systems},
  url = {https://arxiv.org/abs/2412.04453}
}

@inproceedings{jia2026guidedvla,
  title = {{GuidedVLA}: Specifying Task-Relevant Factors via Plug-and-Play Action Attention Specialization},
  author = {Jia, Xiaosong and Yang, Bowen and Ge, Zuhao and Nie, Xian and Zhou, Yuchen and Fan, Cunxin and Li, Yufeng and Chai, Yilin and Jing, Chao and Liang, Zijian and Bu, Qingwen and Cao, Haidong and Wu, Chao and Li, Qifeng and Yang, Zhenjie and Zhang, Chenhe and Li, Hongyang and Wu, Zuxuan and Yan, Junchi and Jiang, Yu-Gang},
  year = {2026},
  booktitle = {Robotics: Science and Systems},
  url = {https://arxiv.org/abs/2605.12369}
}

@inproceedings{wang2025auxthink,
  title = {{Aux-Think}: Exploring Reasoning Strategies for Data-Efficient Vision-Language Navigation},
  author = {Wang, Shuo and Wang, Yongcai and Li, Wanting and Cai, Xudong and Wang, Yucheng and Chen, Maiyue and Wang, Kaihui and Su, Zhizhong and Li, Deying and Fan, Zhaoxin},
  year = {2025},
  booktitle = {Advances in Neural Information Processing Systems},
  url = {https://arxiv.org/abs/2505.11886}
}

@inproceedings{wang2025vggt,
  title = {{VGGT}: Visual Geometry Grounded Transformer},
  author = {Wang, Jianyuan and Chen, Minghao and Karaev, Nikita and Vedaldi, Andrea and Rupprecht, Christian and Novotny, David},
  year = {2025},
  booktitle = {CVPR},
  url = {https://arxiv.org/abs/2503.11651}
}

@article{zhang2026futurenav,
  title = {{FutureNav}: Unified World-Action Modeling for Vision-and-Language Navigation},
  author = {Zhang, Lingfeng and Gong, Zeying and Hao, Xiaoshuai and Fu, Haoxiang and Zhang, Qiang and Zhou, Mingliang and Ye, Hangjun and Liang, Xiaojun and Liang, Junwei and Ding, Wenbo},
  year = {2026},
  journal = {arXiv preprint arXiv:2606.30367},
  url = {https://arxiv.org/abs/2606.30367}
}

@article{bai2025qwen3vl,
  title = {{Qwen3-VL} Technical Report},
  author = {Bai, Shuai and Cai, Yuxuan and Chen, Ruizhe and Chen, Keqin and Chen, Xionghui and others},
  year = {2025},
  journal = {arXiv preprint arXiv:2511.21631},
  url = {https://arxiv.org/abs/2511.21631}
}

@article{ilharco2019ndtw,
  title = {General Evaluation for Instruction Conditioned Navigation using Dynamic Time Warping},
  author = {Ilharco, Gabriel and Jain, Vihan and Ku, Alexander and Ie, Eugene and Baldridge, Jason},
  year = {2019},
  journal = {arXiv preprint arXiv:1907.05446},
  url = {https://arxiv.org/abs/1907.05446}
}

@inproceedings{krantz2021waypoint,
  title     = {Waypoint Models for Instruction-Guided Navigation in Continuous Environments},
  author    = {Krantz, Jacob and Gokaslan, Aaron and Batra, Dhruv and Lee, Stefan and Maksymets, Oleksandr},
  booktitle = {Proceedings of the IEEE/CVF International Conference on Computer Vision},
  pages     = {15162--15171},
  year      = {2021},
  url       = {https://openaccess.thecvf.com/content/ICCV2021/html/Krantz_Waypoint_Models_for_Instruction-Guided_Navigation_in_Continuous_Environments_ICCV_2021_paper.html}
}

@inproceedings{hong2022bridging,
  title     = {Bridging the Gap Between Learning in Discrete and Continuous Environments for Vision-and-Language Navigation},
  author    = {Hong, Yicong and Wang, Zun and Wu, Qi and Gould, Stephen},
  booktitle = {Proceedings of the IEEE/CVF Conference on Computer Vision and Pattern Recognition},
  pages     = {15439--15449},
  year      = {2022},
  url       = {https://openaccess.thecvf.com/content/CVPR2022/html/Hong_Bridging_the_Gap_Between_Learning_in_Discrete_and_Continuous_Environments_CVPR_2022_paper.html}
}

@inproceedings{krantz2022sim2sim,
  title     = {Sim-2-Sim Transfer for Vision-and-Language Navigation in Continuous Environments},
  author    = {Krantz, Jacob and Lee, Stefan},
  booktitle = {European Conference on Computer Vision},
  pages     = {588--603},
  year      = {2022},
  url       = {https://arxiv.org/abs/2204.09667}
}

@inproceedings{hong2023ego2map,
  title     = {Learning Navigational Visual Representations with Semantic Map Supervision},
  author    = {Hong, Yicong and Zhou, Yang and Zhang, Ruiyi and Dernoncourt, Franck and Bui, Trung and Gould, Stephen and Tan, Hao},
  booktitle = {Proceedings of the IEEE/CVF International Conference on Computer Vision},
  pages     = {3055--3067},
  year      = {2023},
  url       = {https://openaccess.thecvf.com/content/ICCV2023/html/Hong_Learning_Navigational_Visual_Representations_with_Semantic_Map_Supervision_ICCV_2023_paper.html}
}

@inproceedings{wang2023dreamwalker,
  title     = {{DREAMWALKER}: Mental Planning for Continuous Vision-Language Navigation},
  author    = {Wang, Hanqing and Liang, Wei and Van Gool, Luc and Wang, Wenguan},
  booktitle = {Proceedings of the IEEE/CVF International Conference on Computer Vision},
  pages     = {10873--10883},
  year      = {2023},
  url       = {https://openaccess.thecvf.com/content/ICCV2023/html/Wang_DREAMWALKER_Mental_Planning_for_Continuous_Vision-Language_Navigation_ICCV_2023_paper.html}
}

@inproceedings{wang2023gridmm,
  title     = {{GridMM}: Grid Memory Map for Vision-and-Language Navigation},
  author    = {Wang, Zihan and Li, Xiangyang and Yang, Jiahao and Liu, Yeqi and Jiang, Shuqiang},
  booktitle = {Proceedings of the IEEE/CVF International Conference on Computer Vision},
  pages     = {15625--15636},
  year      = {2023},
  url       = {https://openaccess.thecvf.com/content/ICCV2023/html/Wang_GridMM_Grid_Memory_Map_for_Vision-and-Language_Navigation_ICCV_2023_paper.html}
}

@article{an2022reborn,
  title   = {1st Place Solutions for {RxR-Habitat} Vision-and-Language Navigation Competition ({CVPR} 2022)},
  author  = {An, Dong and Wang, Zun and Li, Yangguang and Wang, Yi and Hong, Yicong and Huang, Yan and Wang, Liang and Shao, Jing},
  journal = {arXiv preprint arXiv:2206.11610},
  year    = {2022},
  url     = {https://arxiv.org/abs/2206.11610}
}

@inproceedings{long2025instructnav,
  title     = {{InstructNav}: Zero-shot System for Generic Instruction Navigation in Unexplored Environment},
  author    = {Long, Yuxing and Cai, Wenzhe and Wang, Hongcheng and Zhan, Guanqi and Dong, Hao},
  booktitle = {Proceedings of the 8th Conference on Robot Learning},
  series    = {Proceedings of Machine Learning Research},
  volume    = {270},
  pages     = {2049--2060},
  publisher = {PMLR},
  year      = {2025},
  note      = {CoRL 2024},
  url       = {https://proceedings.mlr.press/v270/long25b.html}
}

@inproceedings{zhang2025cosmo,
  title     = {{COSMO}: Combination of Selective Memorization for Low-Cost Vision-and-Language Navigation},
  author    = {Zhang, Siqi and Qiao, Yanyuan and Wang, Qunbo and Yan, Zike and Wu, Qi and Wei, Zhihua and Liu, Jing},
  booktitle = {Proceedings of the IEEE/CVF International Conference on Computer Vision},
  pages     = {5511--5522},
  year      = {2025},
  url       = {https://arxiv.org/abs/2503.24065}
}

@inproceedings{chen2025aoplanner,
  title     = {Affordances-Oriented Planning Using Foundation Models for Continuous Vision-Language Navigation},
  author    = {Chen, Jiaqi and Lin, Bingqian and Liu, Xinmin and Ma, Lin and Liang, Xiaodan and Wong, Kwan-Yee K.},
  booktitle = {Proceedings of the AAAI Conference on Artificial Intelligence},
  volume    = {39},
  pages     = {23568--23576},
  year      = {2025},
  url       = {https://ojs.aaai.org/index.php/AAAI/article/view/34526}
}

@inproceedings{raychaudhuri2021law,
  title     = {Language-Aligned Waypoint ({LAW}) Supervision for Vision-and-Language Navigation in Continuous Environments},
  author    = {Raychaudhuri, Sonia and Wani, Saim and Patel, Shivansh and Jain, Unnat and Chang, Angel},
  booktitle = {Proceedings of the 2021 Conference on Empirical Methods in Natural Language Processing},
  pages     = {4018--4028},
  publisher = {Association for Computational Linguistics},
  year      = {2021},
  url       = {https://aclanthology.org/2021.emnlp-main.328/}
}

@inproceedings{zhang2025mapnav,
  title     = {{MapNav}: A Novel Memory Representation via Annotated Semantic Maps for {VLM}-based Vision-and-Language Navigation},
  author    = {Zhang, Lingfeng and Hao, Xiaoshuai and Xu, Qinwen and Zhang, Qiang and Zhang, Xinyao and Wang, Pengwei and Zhang, Jing and Wang, Zhongyuan and Zhang, Shanghang and Xu, Renjing},
  booktitle = {Proceedings of the 63rd Annual Meeting of the Association for Computational Linguistics (Volume 1: Long Papers)},
  pages     = {13032--13056},
  publisher = {Association for Computational Linguistics},
  year      = {2025},
  url       = {https://aclanthology.org/2025.acl-long.638/}
}

@inproceedings{wang2025g3dlf,
  title     = {{g3D-LF}: Generalizable {3D}-Language Feature Fields for Embodied Tasks},
  author    = {Wang, Zihan and Lee, Gim Hee},
  booktitle = {Proceedings of the IEEE/CVF Conference on Computer Vision and Pattern Recognition},
  pages     = {14191--14202},
  year      = {2025},
  url       = {https://arxiv.org/abs/2411.17030}
}

@inproceedings{liu2025navid4d,
  title     = {{NaVid-4D}: Unleashing Spatial Intelligence in Egocentric {RGB-D} Videos for Vision-and-Language Navigation},
  author    = {Liu, Haoran and Wan, Weikang and Yu, Xiqian and Li, Minghan and Zhang, Jiazhao and Zhao, Bo and Chen, Zhibo and Wang, Zhongyuan and Zhang, Zhizheng and Wang, He},
  booktitle = {2025 IEEE International Conference on Robotics and Automation (ICRA)},
  pages     = {10607--10615},
  year      = {2025},
  url       = {https://lhrrhl0419.github.io/NaVid4D/}
}

@inproceedings{yao2025navmorph,
  title     = {{NavMorph}: A Self-Evolving World Model for Vision-and-Language Navigation in Continuous Environments},
  author    = {Yao, Xuan and Gao, Junyu and Xu, Changsheng},
  booktitle = {Proceedings of the IEEE/CVF International Conference on Computer Vision},
  pages     = {5536--5546},
  year      = {2025},
  url       = {https://openaccess.thecvf.com/content/ICCV2025/html/Yao_NavMorph_A_Self-Evolving_World_Model_for_Vision-and-Language_Navigation_in_Continuous_ICCV_2025_paper.html}
}

@inproceedings{yang2026gavln,
  title     = {{GA-VLN}: Geometry-Aware {BEV} Representation for Efficient Vision-Language Navigation},
  author    = {Yang, Jiahao and Wang, Zihan and Li, Xiangyang and Zhu, Xing and Shen, Yujun and Xu, Yinghao and Jiang, Shuqiang},
  booktitle = {Proceedings of the IEEE/CVF Conference on Computer Vision and Pattern Recognition},
  year      = {2026},
  url       = {https://arxiv.org/abs/2605.22036}
}

@inproceedings{wang2025sim2real,
  title     = {Sim-to-Real Transfer via {3D} Feature Fields for Vision-and-Language Navigation},
  author    = {Wang, Zihan and Li, Xiangyang and Yang, Jiahao and Liu, Yeqi and Jiang, Shuqiang},
  booktitle = {Proceedings of the 8th Conference on Robot Learning},
  series    = {Proceedings of Machine Learning Research},
  volume    = {270},
  pages     = {2982--2995},
  publisher = {PMLR},
  year      = {2025},
  note      = {CoRL 2024},
  url       = {https://proceedings.mlr.press/v270/wang25i.html}
}

@inproceedings{zhang2025uninavid,
  title     = {{Uni-NaVid}: A Video-based Vision-Language-Action Model for Unifying Embodied Navigation Tasks},
  author    = {Zhang, Jiazhao and Wang, Kunyu and Wang, Shaoan and Li, Minghan and Liu, Haoran and Wei, Songlin and Wang, Zhongyuan and Zhang, Zhizheng and Wang, He},
  booktitle = {Robotics: Science and Systems},
  year      = {2025},
  doi       = {10.15607/RSS.2025.XXI.013},
  url       = {https://www.roboticsproceedings.org/rss21/p013.html}
}

@techreport{internrobotics2025internvlan1,
  title       = {{InternVLA-N1}: An Open Dual-System Vision-Language Navigation Foundation Model with Learned Latent Plans},
  author      = {{Intern Robotics}},
  institution = {Shanghai Artificial Intelligence Laboratory},
  year        = {2025},
  url         = {https://internrobotics.github.io/internvla-n1.github.io/static/pdfs/InternVLA_N1.pdf}
}

@inproceedings{wang2026monodream,
  title     = {{MonoDream}: Monocular Vision-Language Navigation with Panoramic Dreaming},
  author    = {Wang, Shuo and Wang, Yongcai and Fan, Zhaoxin and Wang, Yucheng and Chen, Maiyue and Wang, Kaihui and Su, Zhizhong and Li, Wanting and Cai, Xudong and Jin, Yeying and Li, Deying},
  booktitle = {Proceedings of the AAAI Conference on Artificial Intelligence},
  volume    = {40},
  pages     = {10074--10082},
  year      = {2026},
  doi       = {10.1609/aaai.v40i12.37974},
  url       = {https://ojs.aaai.org/index.php/AAAI/article/view/37974}
}

@inproceedings{zhang2026navfom,
  title     = {Embodied Navigation Foundation Model},
  author    = {Zhang, Jiazhao and Li, Anqi and Qi, Yunpeng and Li, Minghan and Liu, Jiahang and Wang, Shaoan and Liu, Haoran and Zhou, Gengze and Wu, Yuze and Li, Xingxing and Fan, Yuxin and Li, Wenjun and Chen, Zhibo and Gao, Fei and Wu, Qi and Zhang, Zhizheng and Wang, He},
  booktitle = {International Conference on Learning Representations},
  year      = {2026},
  url       = {https://openreview.net/forum?id=kkBOIsrCXh}
}

@inproceedings{chen2022duet,
  title = {Think Global, Act Local: Dual-Scale Graph Transformer for Vision-and-Language Navigation},
  author = {Chen, Shizhe and Guhur, Pierre-Louis and Tapaswi, Makarand and Schmid, Cordelia and Laptev, Ivan},
  booktitle = {Proceedings of the IEEE/CVF Conference on Computer Vision and Pattern Recognition},
  pages = {16537--16547},
  year = {2022},
  url = {https://openaccess.thecvf.com/content/CVPR2022/html/Chen_Think_Global_Act_Local_Dual-Scale_Graph_Transformer_for_Vision-and-Language_Navigation_CVPR_2022_paper.html}
}

@article{an2024etpnav,
  title = {{ETPNav}: Evolving Topological Planning for Vision-Language Navigation in Continuous Environments},
  author = {An, Dong and Wang, Hanqing and Wang, Wenguan and Wang, Zun and Huang, Yan and He, Keji and Wang, Liang},
  journal = {IEEE Transactions on Pattern Analysis and Machine Intelligence},
  year = {2024},
  url = {https://github.com/MarSaKi/ETPNav}
}

@article{lin2025navcot,
  title = {{NavCoT}: Boosting {LLM}-Based Vision-and-Language Navigation via Learning Disentangled Reasoning},
  author = {Lin, Bingqian and Nie, Yunshuang and Wei, Ziming and Chen, Jiaqi and Ma, Shikui and Han, Jianhua and Xu, Hang and Chang, Xiaojun and Liang, Xiaodan},
  journal = {IEEE Transactions on Pattern Analysis and Machine Intelligence},
  year = {2025},
  url = {https://github.com/expectorlin/NavCoT}
}

@inproceedings{wang2026progressthink,
  title = {{Progress-Think}: Semantic Progress Reasoning for Vision-Language Navigation},
  author = {Wang, Shuo and Wang, Yucheng and Lian, Guoxin and Wang, Yongcai and Chen, Maiyue and Wang, Kaihui and Zhang, Bo and Su, Zhizhong and Zhou, Yutian and Li, Wanting and Li, Deying and Fan, Zhaoxin},
  booktitle = {Proceedings of the IEEE/CVF Conference on Computer Vision and Pattern Recognition},
  pages = {4076--4086},
  year = {2026},
  url = {https://openaccess.thecvf.com/content/CVPR2026/html/Wang_Progress-Think_Semantic_Progress_Reasoning_for_Vision-Language_Navigation_CVPR_2026_paper.html}
}

@inproceedings{guo2026awarevln,
  title = {{AwareVLN}: Reasoning with Self-awareness for Vision-Language Navigation},
  author = {Guo, Wenxuan and Xu, Xiuwei and Liu, Yichen and Li, Xiangyu and Yin, Hang and Chen, Huangxing and Zheng, Wenzhao and Feng, Jianjiang and Zhou, Jie and Lu, Jiwen},
  booktitle = {Proceedings of the IEEE/CVF Conference on Computer Vision and Pattern Recognition},
  pages = {4065--4075},
  year = {2026},
  url = {https://openaccess.thecvf.com/content/CVPR2026/html/Guo_AwareVLN_Reasoning_with_Self-awareness_for_Vision-Language_Navigation_CVPR_2026_paper.html}
}

@inproceedings{li2023kerm,
  title = {{KERM}: Knowledge Enhanced Reasoning for Vision-and-Language Navigation},
  author = {Li, Xiangyang and Wang, Zihan and Yang, Jiahao and Wang, Yaowei and Jiang, Shuqiang},
  booktitle = {Proceedings of the IEEE/CVF Conference on Computer Vision and Pattern Recognition},
  pages = {2583--2592},
  year = {2023},
  url = {https://openaccess.thecvf.com/content/CVPR2023/html/Li_KERM_Knowledge_Enhanced_Reasoning_for_Vision-and-Language_Navigation_CVPR_2023_paper.html}
}

@inproceedings{wei2026streamvln,
  title = {{StreamVLN}: Streaming Vision-and-Language Navigation via {SlowFast} Context Modeling},
  author = {Wei, Meng and Wan, Chenyang and Yu, Xiqian and Wang, Tai and Yang, Yuqiang and Mao, Xiaohan and Zhu, Chenming and Cai, Wenzhe and Wang, Hanqing and Chen, Yilun and Liu, Xihui and Pang, Jiangmiao},
  booktitle = {IEEE International Conference on Robotics and Automation},
  year = {2026},
  url = {https://github.com/InternRobotics/StreamVLN}
}

@inproceedings{li2026spatialforcing,
  title = {Spatial Forcing: Implicit Spatial Representation Alignment for Vision-Language-Action Model},
  author = {Li, Fuhao and Song, Wenxuan and Zhao, Han and Wang, Jingbo and Ding, Pengxiang and Wang, Donglin and Zeng, Long and Li, Haoang},
  booktitle = {International Conference on Learning Representations},
  year = {2026},
  url = {https://openreview.net/forum?id=euMVC1DO4k}
}

@inproceedings{anderson2018vln,
  title = {Vision-and-Language Navigation: Interpreting Visually-Grounded Navigation Instructions in Real Environments},
  author = {Anderson, Peter and Wu, Qi and Teney, Damien and Bruce, Jake and Johnson, Mark and S{\"u}nderhauf, Niko and Reid, Ian and Gould, Stephen and van den Hengel, Anton},
  booktitle = {Proceedings of the IEEE Conference on Computer Vision and Pattern Recognition},
  pages = {3674--3683},
  year = {2018},
  url = {https://openaccess.thecvf.com/content_cvpr_2018/html/Anderson_Vision-and-Language_Navigation_Interpreting_CVPR_2018_paper.html}
}

@inproceedings{an2023bevbert,
  title = {{BEVBert}: Multimodal Map Pre-training for Language-guided Navigation},
  author = {An, Dong and Qi, Yuankai and Li, Yangguang and Huang, Yan and Wang, Liang and Tan, Tieniu and Shao, Jing},
  booktitle = {Proceedings of the IEEE/CVF International Conference on Computer Vision},
  year = {2023},
  url = {https://github.com/MarSaKi/VLN-BEVBert}
}

@inproceedings{zhou2024navgpt,
  title = {{NavGPT}: Explicit Reasoning in Vision-and-Language Navigation with Large Language Models},
  author = {Zhou, Gengze and Hong, Yicong and Wu, Qi},
  booktitle = {Proceedings of the AAAI Conference on Artificial Intelligence},
  volume = {38},
  pages = {7641--7649},
  year = {2024},
  doi = {10.1609/aaai.v38i7.28597},
  url = {https://ojs.aaai.org/index.php/AAAI/article/view/28597}
}

@inproceedings{cui2023gela,
  title = {Grounded Entity-Landmark Adaptive Pre-Training for Vision-and-Language Navigation},
  author = {Cui, Yibo and Xie, Liang and Zhang, Yakun and Zhang, Meishan and Yan, Ye and Yin, Erwei},
  booktitle = {Proceedings of the IEEE/CVF International Conference on Computer Vision},
  pages = {12043--12053},
  year = {2023},
  url = {https://openaccess.thecvf.com/content/ICCV2023/html/Cui_Grounded_Entity-Landmark_Adaptive_Pre-Training_for_Vision-and-Language_Navigation_ICCV_2023_paper.html}
}

@inproceedings{hong2020subinstruction,
  title = {Sub-Instruction Aware Vision-and-Language Navigation},
  author = {Hong, Yicong and Rodriguez, Cristian and Wu, Qi and Gould, Stephen},
  booktitle = {Proceedings of the 2020 Conference on Empirical Methods in Natural Language Processing},
  pages = {3360--3376},
  year = {2020},
  doi = {10.18653/v1/2020.emnlp-main.271},
  url = {https://aclanthology.org/2020.emnlp-main.271/}
}

@inproceedings{liu2024groundingdino,
  title = {Grounding {DINO}: Marrying {DINO} with Grounded Pre-Training for Open-Set Object Detection},
  author = {Liu, Shilong and Zeng, Zhaoyang and Ren, Tianhe and Li, Feng and Zhang, Hao and Yang, Jie and Jiang, Qing and Li, Chunyuan and Yang, Jianwei and Su, Hang and Zhu, Jun and Zhang, Lei},
  booktitle = {Proceedings of the European Conference on Computer Vision},
  year = {2024},
  url = {https://www.ecva.net/papers/eccv_2024/papers_ECCV/papers/06319.pdf}
}

@inproceedings{ravi2025sam2,
  title = {{SAM 2}: Segment Anything in Images and Videos},
  author = {Ravi, Nikhila and Gabeur, Valentin and Hu, Yuan-Ting and Hu, Ronghang and Ryali, Chaitanya and Ma, Tengyu and Khedr, Haitham and R{\"a}dle, Roman and Rolland, Chloe and Gustafson, Laura and Mintun, Eric and Pan, Junting and Alwala, Kalyan Vasudev and Carion, Nicolas and Wu, Chao-Yuan and Girshick, Ross and Doll{\'a}r, Piotr and Feichtenhofer, Christoph},
  booktitle = {International Conference on Learning Representations},
  year = {2025},
  url = {https://proceedings.iclr.cc/paper_files/paper/2025/file/45c1f6a8cbf2da59ebf2c802b4f742cd-Paper-Conference.pdf}
}

@inproceedings{ross2011dagger,
  title = {A Reduction of Imitation Learning and Structured Prediction to No-Regret Online Learning},
  author = {Ross, Stephane and Gordon, Geoffrey and Bagnell, Drew},
  booktitle = {Proceedings of the Fourteenth International Conference on Artificial Intelligence and Statistics},
  volume = {15},
  pages = {627--635},
  year = {2011},
  url = {https://proceedings.mlr.press/v15/ross11a.html}
}

@article{zhang2025llavavideo,
  title = {{LLaVA-Video}: Video Instruction Tuning With Synthetic Data},
  author = {Zhang, Yuanhan and Wu, Jinming and Li, Wei and Li, Bo and Ma, Zejun and Liu, Ziwei and Li, Chunyuan},
  journal = {Transactions on Machine Learning Research},
  year = {2025},
  url = {https://arxiv.org/abs/2410.02713}
}

@article{yang2025cambrians,
  title = {{Cambrian-S}: Towards Spatial Supersensing in Video},
  author = {Yang, Shusheng and Yang, Jihan and Huang, Pinzhi and Brown, Ellis and Yang, Zihao and Yu, Yue and Tong, Shengbang and Zheng, Zihan and Xu, Yifan and Wang, Muhan and Lu, Daohan and Fergus, Rob and LeCun, Yann and Fei-Fei, Li and Xie, Saining},
  journal = {arXiv preprint arXiv:2511.04670},
  year = {2025},
  url = {https://arxiv.org/abs/2511.04670}
}

@inproceedings{davis2006prroc,
  title = {The Relationship Between Precision-Recall and ROC Curves},
  author = {Davis, Jesse and Goadrich, Mark},
  booktitle = {Proceedings of the International Conference on Machine Learning},
  pages = {233--240},
  year = {2006},
  doi = {10.1145/1143844.1143874}
}

@inproceedings{zhang2016excitation,
  title = {Top-down Neural Attention by Excitation Backprop},
  author = {Zhang, Jianming and Lin, Zhe and Brandt, Jonathan and Shen, Xiaohui and Sclaroff, Stan},
  booktitle = {European Conference on Computer Vision},
  year = {2016},
  url = {https://arxiv.org/abs/1608.00507}
}

@inproceedings{wang2020scorecam,
  title = {{Score-CAM}: Score-Weighted Visual Explanations for Convolutional Neural Networks},
  author = {Wang, Haofan and Wang, Zifan and Du, Mengnan and Yang, Fan and Zhang, Zijian and Ding, Sirui and Mardziel, Piotr and Hu, Xia},
  booktitle = {Proceedings of the IEEE/CVF Conference on Computer Vision and Pattern Recognition Workshops},
  year = {2020},
  url = {https://arxiv.org/abs/1910.01279}
}

@inproceedings{eigen2014depth,
  title = {Depth Map Prediction from a Single Image using a Multi-Scale Deep Network},
  author = {Eigen, David and Puhrsch, Christian and Fergus, Rob},
  booktitle = {Advances in Neural Information Processing Systems},
  volume = {27},
  year = {2014},
  url = {https://arxiv.org/abs/1406.2283}
}

@inproceedings{chen2016depthwild,
  title = {Single-Image Depth Perception in the Wild},
  author = {Chen, Weifeng and Fu, Zhao and Yang, Dawei and Deng, Jia},
  booktitle = {Advances in Neural Information Processing Systems},
  volume = {29},
  year = {2016},
  url = {https://papers.nips.cc/paper_files/paper/2016/hash/0deb1c54814305ca9ad266f53bc82511-Abstract.html}
}

@inproceedings{ma2019selfmonitoring,
 title={Self-Monitoring Navigation Agent via Auxiliary Progress Estimation},
 author={Ma, Chih-Yao and Lu, Jiasen and Wu, Zuxuan and AlRegib, Ghassan and Kira, Zsolt and Socher, Richard and Xiong, Caiming},
 booktitle={International Conference on Learning Representations},
 year={2019},
 url={https://arxiv.org/abs/1901.03035}
}

@inproceedings{zhu2020auxrn,
 title={Vision-Language Navigation With Self-Supervised Auxiliary Reasoning Tasks},
 author={Zhu, Fengda and Zhu, Yi and Chang, Xiaojun and Liang, Xiaodan},
 booktitle={Proceedings of the IEEE/CVF Conference on Computer Vision and Pattern Recognition},
 year={2020},
 url={https://openaccess.thecvf.com/content_CVPR_2020/papers/Zhu_Vision-Language_Navigation_With_Self-Supervised_Auxiliary_Reasoning_Tasks_CVPR_2020_paper.pdf}
}

@inproceedings{he2021landmarkrxr,
 title={{Landmark-RxR}: Solving Vision-and-Language Navigation with Fine-Grained Alignment Supervision},
 author={He, Keji and Huang, Yan and Wu, Qi and Yang, Jianhua and An, Dong and Sima, Shuanglin and Wang, Liang},
 booktitle={Advances in Neural Information Processing Systems},
 volume={34},
 year={2021},
 url={https://proceedings.neurips.cc/paper/2021/hash/0602940f23884f782058efac46f64b0f-Abstract.html}
}

@inproceedings{selvaraju2019hint,
 title={Taking a {HINT}: Leveraging Explanations to Make Vision and Language Models More Grounded},
 author={Selvaraju, Ramprasaath R. and Lee, Stefan and Shen, Yilin and Jin, Hongxia and Ghosh, Shalini and Heck, Larry and Batra, Dhruv and Parikh, Devi},
 booktitle={Proceedings of the IEEE/CVF International Conference on Computer Vision},
 year={2019},
 url={https://openaccess.thecvf.com/content_ICCV_2019/html/Selvaraju_Taking_a_HINT_Leveraging_Explanations_to_Make_Vision_and_Language_ICCV_2019_paper.html}
}
